\documentclass{article} 
\usepackage[a4paper]{geometry}
\usepackage{url}  
\usepackage{graphicx}  
\usepackage[numbers,sort&compress]{natbib}
\usepackage{enumitem}
\usepackage{amsfonts}
\usepackage[utf8]{inputenc}
\usepackage{color}
\usepackage{parskip}
\usepackage{bm}
\usepackage{amsmath}
\usepackage{amssymb}
\usepackage{mathtools}
\usepackage{algorithm2e}
\usepackage{xcolor}
\usepackage{multirow}
\usepackage{subfig}
\usepackage{authblk}
\usepackage{hyperref}
\usepackage{array}
\usepackage{pifont}

\ifdefined\nohyperref\else\ifdefined\hypersetup
\definecolor{mydarkblue}{rgb}{0,0.08,0.45}
\definecolor{mydarkred}{rgb}{0.64,0,0}
\hypersetup{ %
	pdftitle={},
	pdfauthor={},
	pdfkeywords={},
	pdfborder=0 0 0,
	pdfpagemode=UseNone,
	colorlinks=true,
	linkcolor=mydarkblue,
	citecolor=mydarkred,
	filecolor=cyan,
	urlcolor=magenta,
	pdfview=FitH}

\SetKwInput{KwInput}{Input}                
\SetKwInput{KwOutput}{Output}              

\SetCommentSty{mycommfont}

\makeatletter
\newcommand{\removelatexerror}{\let\@latex@error\@gobble}
\makeatother

\makeatletter
\newcommand{\printfnsymbol}[1]{%
	\textsuperscript{\@fnsymbol{#1}}%
}

\title{A Decision-Theoretic Approach for Model Interpretability in Bayesian Framework}

\author[1]{Homayun Afrabandpey}
\author[1,2,\ding{169}]{Tomi Peltola}
\author[2,\ding{168}]{Juho Piironen} 
\author[1]{Aki Vehtari}
\author[1,3]{Samuel Kaski}

\affil[1]{Helsinki Institute for Information Technology HIIT,\protect\\ Department of Computer Science, Aalto University}
\affil[2]{Curious AI}
\affil[3]{University of Manchester, UK}
\makeatletter
\renewcommand\AB@affilsepx{, \protect\Affilfont}
\makeatother
\affil[1]{\texttt{firstname.lastname@aalto.fi}}\affil[\ding{169}]{\texttt{tomi.peltola@tmpl.fi}}\affil[\ding{168}]{\texttt{juho.t.piironen@gmail.com}}
\date{}

\begin{document}
	\maketitle
	\begin{abstract}
		A salient approach to interpretable machine learning is to restrict modeling to simple models. In the Bayesian framework, this can be pursued by restricting the model structure and prior to favor interpretable models. Fundamentally, however, interpretability is about users' preferences, not the data generation mechanism; it is more natural to formulate interpretability as a utility function. In this work, we propose an interpretability utility, which explicates the trade-off between explanation fidelity and interpretability in the Bayesian framework. The method consists of two steps. First, a reference model, possibly a black-box Bayesian predictive model which does not compromise accuracy, is fitted to the training data. Second, a proxy model from an interpretable model family that best mimics the predictive behaviour of the reference model is found by optimizing the interpretability utility function. The approach is model agnostic -- neither the interpretable model nor the reference model are restricted to a certain class of models -- and the optimization problem can be solved using standard tools. Through experiments on real-word data sets, using decision trees as interpretable models and Bayesian additive regression models as reference models, we show that for the same level of interpretability, our approach generates more accurate models than the alternative of restricting the prior. We also propose a systematic way to measure stability of interpretabile models constructed by different interpretability approaches and show that our proposed approach generates more stable models.
	\end{abstract}
	
	\section{Introduction and Background}\label{Sec:intro}
	
	Accurate machine learning (ML) models are usually complex and opaque, even to the modelers who built them \cite{lipton2018mythos}. This lack of interpretability remains a key barrier to the adoption of ML models in some applications including health care and economy. To bridge this gap, there is growing interest among the ML community to interpretability methods. 
	Such methods can be divided into (i) interpretable model construction, and (ii) post-hoc interpretation. The former aims at constructing models that are understandable. Post-hoc interpretation approaches can be categorized further into (i) model-level interpretation (a.k.a. global interpretation), and (ii) prediction-level interpretation (a.k.a. local interpretation) \cite{du2018techniques}. Model-level interpretation aims at making existing black-box models interpretable. Prediction-level interpretation aims at explaining each individual prediction made by the model \cite{doshi2017towards}. In this paper, we focus mostly on post-hoc interpretation. 
	
	Prior research on the construction of interpretable models has mainly focused on restricting modeling to simple and easy-to-understand models. Examples of such models include sparse linear models \cite{ustun2016supersparse}, generalized additive models \cite{lou2012intelligible}, decision sets \cite{lakkaraju2016interpretable}, and rule lists \cite{jung2017simple}. In the Bayesian framework, this approach maps to defining model structure and prior distributions that favor interpretable models \cite{letham2015interpretable, wang2017bayesian, popkes2019interpretable, wang2018multi}. We call this approach \textit{interpretability prior}. Letham et al.~\cite{letham2015interpretable} established an interpretability prior approach for classification by use of decision lists. Interpretability measures used to define the priors were (i) the number of rules in the list and (ii) the size of the rules (number of statements in the left-hand side of rules). A prior distribution was defined over rule lists to favor decision lists with a small number of short rules. Wang et al.~\cite{wang2017bayesian} developed two probabilistic models for interpretable classification by constructing rule sets in the form of Disjunctive Normal Forms (DNFs). In this work, interpretability is achieved similar to \cite{letham2015interpretable}, using prior distributions which favor rule sets with a smaller number of short rules. In \cite{wang2018multi}, the authors extended \cite{wang2017bayesian} by presenting a multi-value rule set for interpretable classification, which allows multiple values per condition and thereby induces more concise rules compared to single-value rules. As in \cite{wang2017bayesian}, interpretability is characterized by a prior distribution that favors a smaller number of short rules. Popkes et al.~\cite{popkes2019interpretable} built up an interpretable Bayesian neural network for clinical decision-making tasks, where interpretability is attained by employing a sparsity-inducing prior over feature weights. For more examples, see \cite{kim2015ibcm,hara2016making,yang2017scalable,guo2017bayesian}.
	
	A common practice in model-level interpretability is to use simple models as interpretable surrogates to highly predictive black-box models \cite{craven1996extracting,zhou2016interpreting,bastani2018interpreting,lakkaraju2019faithful,kuttichira2019explaining}. Craven and Shavlik \cite{craven1996extracting} were among the first to adopt this approach for explaining neural networks. They used decision trees as surrogates and trained them to approximate predictions of a neural network. In \cite{zhou2016interpreting}, the authors presented an approach to approximate the predictive behavior of a random forest by use of a single decision tree. With the same objective as \cite{zhou2016interpreting}, Bastani et al.~\cite{bastani2018interpreting} developed an approach to interpret random forests using simple decision trees as surrogates. They employed active learning to construct more accurate decision trees with help from a human. Lakkaraju et al.~\cite{lakkaraju2019faithful} established an approach to interpret black-box classifiers by highlighting the behavior of the black-box model in subspaces characterized by features of user interest. In \cite{kuttichira2019explaining}, the authors used decision trees to extract rules to describe the decision-making behavior of black-box models. For more examples of this approach, see \cite{breiman1996born,meinshausen2010node,wu2018beyond,deng2019interpreting}. The common characteristic of these approaches is that they seek an optimal trade-off between interpretability of the surrogate model and its faithfulness to the black-box model. To the best of our knowledge, there is no Bayesian counterpart for this approach in the interpretability literature.	

	We argue that an interpretability prior is not the best way to optimize interpretability in the Bayesian framework for the following reasons:
	\begin{itemize}
		\item[1.] Interpretability is about users' preferences, not about our assumptions about the data. The prior is meant for the latter. One should distinguish the data generation mechanism from the decision-making process, which in this case includes optimization of interpretability.
		\item[2.] Optimizing interpretability may sacrifice some of the accuracy of the model. If interpretability is pursued by revising the prior, there is no reason why the trade-off between accuracy and interpretability would be optimal. This has been shown for a different but related scenario in \cite{piironen2018projective} where the authors showed that fitting a model using sparsity-inducing priors that favor simpler models results in performance loss.
		\item[3.] Formulating an interpretability prior for certain classes of models such as neural networks could be difficult.
	\end{itemize}
	To address these concerns, we develop a general principle for interpretability in the Bayesian framework, formalizing the idea of approximating black-box models with interpretable surrogates. The approach can be used to both constructing, from scratch, interpretable Bayesian predictive models, or to interpreting existing black-box Bayesian predictive models. The approach consists of two steps: first, a highly accurate Bayesian predictive model, called a reference model, is fitted to the training data without compromising the accuracy. In the second step, an interpretable surrogate model is constructed which best describes locally or globally the behavior of the reference model. The proxy model is obtained by optimizing a utility function, referred to as \textit{interpretability utility}, which consists of two terms: (i) a term to minimize the discrepancy of the proxy model from the reference model, and (ii) a term to penalize the complexity of the model to make the proxy model as interpretable as possible. Term (i) corresponds to selection of reference predictive model in the Bayesian framework \cite[Section~3.3]{vehtari2012survey}. 
	
	The proposed approach can be used both for constructing interpretable Bayesian predictive models and to generate post-hoc interpretation for black-box Bayesian predictive models. When using the approach for post-hoc interpretability, it can be used to generate both global or local interpretation. The approach is model-agnostic, meaning that neither the reference model nor the interpretable proxy are constrained to a particular model family. However, when using the approach to construct interpretable Bayesian predictive models, the surrogate model should be from the family of Bayesian predictive models. We also emphasize that the proposed approach is feasible for non-Bayesian models as well, which can be interpreted to produce point estimates of the parameters of the model instead of posterior distributions. Table \ref{T:comp} compares the characteristics of the proposed approach with some of the related works from literature.
	
	We demonstrate with experiments on real-world data sets that the proposed approach generates more accurate and more stable interpretable models than the alternative of fitting an a priori interpretable model to the data, i.e., using the interpretability prior approach. For the experiments in this paper, decision trees and logistic regression were used as interpretable proxies, and Bayesian additive regression tree (BART) models \cite{chipman2010bart}, Bayesian neural networks, and Gaussian Processes (GP) were used as reference models.
	\begin{table}[!]
		\centering
		\caption{Characteristics of different intereptation approaches. G -- global, L -- local, DT -- Decision Tree, DR -- Decision Rules, M/A -- Model Agnostic, TE -- Tree Ensemble, NN -- Neural Network, C -- Classification, R -- Regression}
		\begin{tabular}{c|c|c|c|c|c|c}
			\hline
			Approach & Ref. & Domain & Interp. Model & Black-Box Model & Task & Bayesian \\ \hline
			Trepan & \cite{craven1996extracting} & G & DT & NN & C & \ding{56} \\ \hline
			-- & \cite{bastani2018interpreting} & G & DT & TE & C & \ding{56} \\ \hline
			BATrees & \cite{breiman1996born} & G & DT & TE & C/R & \ding{56} \\ \hline
			inTrees & \cite{deng2019interpreting} & G & DR & TE & C & \ding{56} \\ \hline
			-- & \cite{kuttichira2019explaining} & G & M/A & M/A & C/R & \ding{56} \\ \hline
			Node Harvest & \cite{meinshausen2010node} & G & TE & TE & R & \ding{56} \\ \hline
			Our Approach & -- & G/L & M/A & M/A & C/R & \ding{52} \\ \hline
		\end{tabular}
		\label{T:comp}
	\end{table}
	
	\subsection{Our Contributions:}
	Main contributions of this paper are:
	\begin{itemize}
		\item We propose a principle for interpretable Bayesian predictive modeling. It combines a reference model with interpretability utility to produce more interpretable models in a decision-theoretically justified way. The proposed approach is model agnostic and can be used with different notions of interpretability.
		\item For the special case of classification and regression tree (CART) \cite{breiman1984classification} as interpretable models and BART as the black-box Bayesian predictive model, we show that the proposed approach outperforms the earlier interpretability prior approach in accuracy, explicating the trade-off between explanation fidelity and interpretability. Further, through experiments with different reference models, i.e., GP and BART, we demonstrate that the predictive power of the reference model positively affects the accuracy of the interpretable model. We also demonstrate that our proposed approach can find a better trade-off between accuracy and interpretability when compared to its non-Bayesian counterparts, i.e., BATrees \cite{breiman1996born} and node harvest \cite{meinshausen2010node}. 
		\item We propose a systematic approach to compare stability of interpretable models and show that the proposed method produces more stable models. 
	\end{itemize}
	\section{Motivation}
	In this section, we discuss the motivation for formulating interpretability optimization in the Bayesian framework as a utility function. We also discuss how this formulation allows to account for model uncertainty in the explanation. Both discussions are accompanied with illustrative examples.
	
	\subsection{Interpretability as a Decision-Making Problem}
	
	Bayesian modeling allows encoding prior information into the prior probability distribution (similarly, one might use regularization in maximum likelihood based inference). This might be tempting to change the prior distribution to favor models that are easier for humans to understand, as has been done in earlier works, using some measure of interpretability. A simple example is to use shrinkage priors in linear regression to find a smaller set of practically important covariates. However, we argue that based on the observation, interpretability is not an inherent characteristic of data generation processes. The approach can be misleading and results in leaking user preferences about interpretability into the model of the data generation process.
	
	We suggest to separate the construction of a model for the data generating process from construction of an interpretable proxy model. In a prediction task, the former corresponds to building a model that predicts as accurately as possible, without restricting it to be interpretable. Interpretability is introduced in the second stage by building an interpretable proxy to explain the behavior of the predictive model. We consider the second step as a decision-making problem, where the task is to choose a proxy model that trades off between human interpretability and fidelity (w.r.t.\ the original model).
	
	
	\subsection{The Issue with Interpretability in the Prior}
	
	Let $\mathcal{M}$ denote the assumptions about the data generating process and $\mathcal{I}$ the preferences toward interpretability. Consider an observation model for data $y$, $p(y\!\mid\!\theta, \mathcal{M})$, and alternative prior distributions $p(\theta\!\mid\!\mathcal{M})$ and $p(\theta\!\mid\!\mathcal{M}, \mathcal{I})$. Here, $\theta$ can, for example, be continuous model parameters (e.g., weights in a regression or classification model) or it can index a set of alternative models (e.g., each configuration of $\theta$ could correspond to using some subset of input variables in a predictive model). Clearly, the posterior distributions $p(\theta\!\mid\!\mathcal{D}, \mathcal{M})$ and $p(\theta\!\mid\!\mathcal{D}, \mathcal{M}, \mathcal{I})$ (and their corresponding posterior predictive distributions) are in general different and the latter includes a bias towards interpretable models. In particular, when $\mathcal{I}$ does not correspond to prior information about the data generation process, there is no guarantee that $p(\theta\!\mid\!\mathcal{D}, \mathcal{M}, \mathcal{I})$ provides a reasonable quantification of our knowledge of $\theta$ given the observations $\mathcal{D}$, or that, $p(\tilde{y}\!\mid\!\mathcal{D}, \mathcal{M}, \mathcal{I})$ provides good predictions. We will give an example of this below. In the special case, where $\mathcal{I}$ does describe the data generation process, it can directly be included in $\mathcal{M}$.
	
	Lage et al~\cite{lage2018human} propose to find interpretable models in two steps: (1) fit a set of models to data and take ones that give high enough predictive accuracy, (2) build a prior over these models, based on an indirect measure of user interpretability (human interpretability score). In practice, the process requires the set of models for step 1 to contain interpretable models, which means that there is still the possibility of leaking user preferences for interpretability into the knowledge about the data generation process. This may lead to an unreasonable trade-off between accuracy and interpretability.
	
	
	
	
	
	\begin{figure}[t!]
		\centering
		\begin{minipage}{0.5\textwidth}    \centering
			\includegraphics[width=\textwidth]{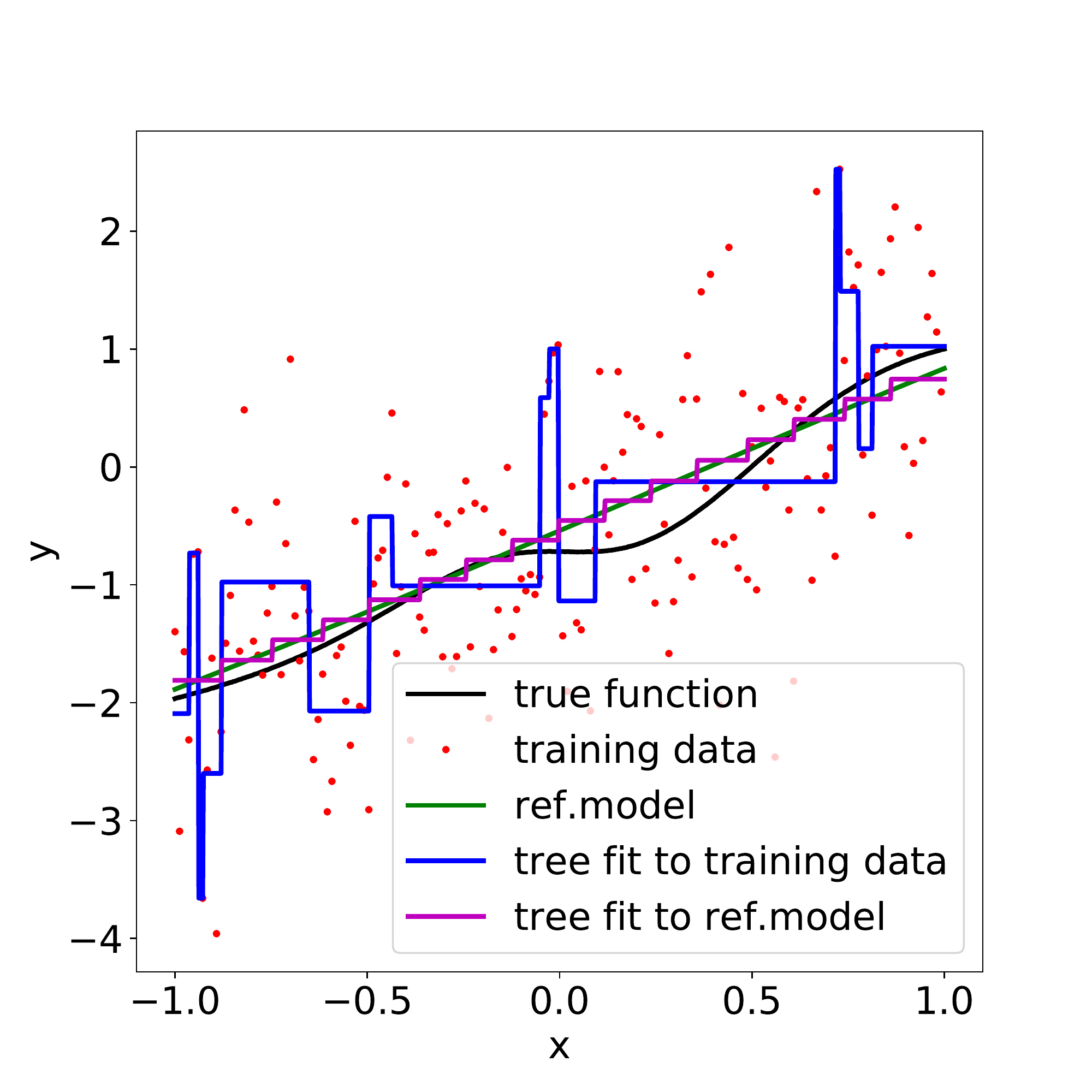}
		\end{minipage}\hspace{-10pt}
		\begin{minipage}{0.5\textwidth}    \centering
			\includegraphics[width=\textwidth]{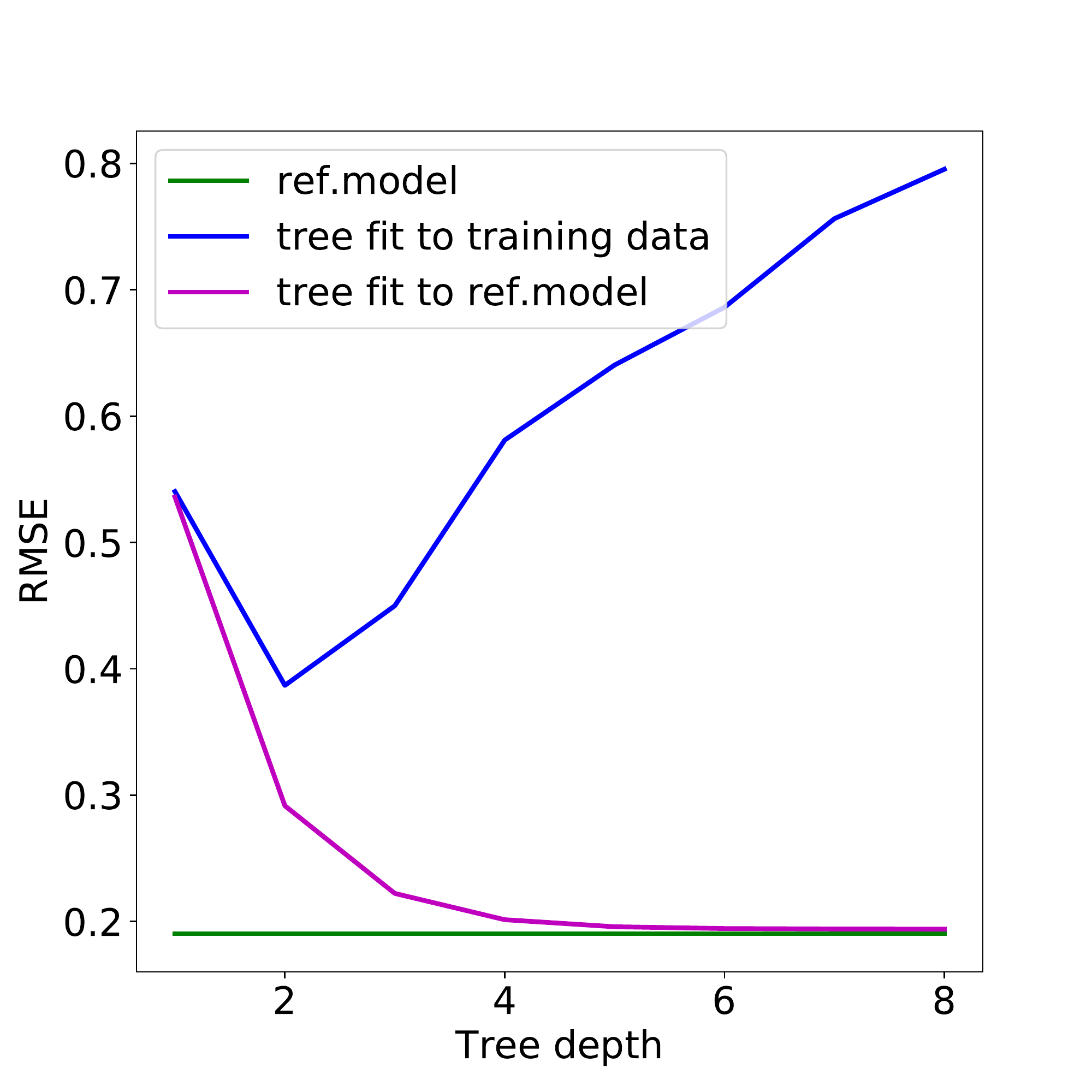}
		\end{minipage}
		\caption{ \textbf{Left}: The reference model (green) is a highly predictive non-interpretable model that approximates the true function (black) well. The interpretable model fitted to the reference model (magenta) approximates the reference model (and consequently the true function) well, while the interpretable model fitted to the training data (blue) fails to approximate the predictive behavior of the true function. \textbf{Right}: Root Mean Squared Errors (RMSE) compared to the true underlying function as the tree depth is varied. By increasing the complexity of the interpretable model (decreasing its interpretability), predictive performance of the reference model and its corresponding interpretable model converge; the interpretable model overfits to the reference model.}
		\label{fig:illustrative_example}
	\end{figure}
	\subsubsection{Illustrative Example}\label{Sec:ref_illust_expl}
	
	We give an example to illustrate the effect of adding interpretability constraints to the prior distribution when these constraints do not match data generating process. For simplicity, we define a single interpretability constraint which is over the structure of the model: regression tree with a fixed depth of 4. The interpretability prior approach corresponds to fitting an interpretable model with the above constraint directly to the training data. In the alternative approach, first a reference model is fitted to the data, and then the reference model is approximated with a proxy model that satisfies the interpretability constraint, using the interpretability utility introduced in Section \ref{Sec:method}. For simplicity of visualization, we use a one-dimensional smooth function as the data-generating process, with Gaussian noise added to observations (Figure~\ref{fig:illustrative_example}:left, black curve and red dots). Regression tree is a piece-wise constant function which does not correspond to the true prior knowledge about the ground-truth function, i.e. being a 1D smooth function. A Gaussian process with the MLP kernel function is used as a reference model for the two-stage approach (Figure~\ref{fig:illustrative_example}:left, magenta).
	
	The regression tree of depth $4$ fitted directly to the data (blue line) overfits and does not give an accurate representation of the underlying data generation process (black line). The two-stage approach, on the other hand, gives a clearly better representation of the smooth, increasing function. This is because the reference model (green line) captures the smoothness of the underlying data generation process and this is transferred to the regression tree (magenta line). The choice of the complexity of the interpretable model is also easier because the tree can only ``overfit'' to the reference model, meaning that it becomes a more accurate (but possibly less easy to interpret) representation of the reference model as shown in Figure~\ref{fig:illustrative_example}:right. 

	\subsection{Interpreting Uncertainty}
	
	In many applications, such as medical treatment effectiveness prediction \cite{sundin2018improving}, knowing the uncertainty in the prediction is important. Any explanation of the predictive model should also provide insight about the uncertainties and their sources. The posterior predictive distribution of the reference model contains both the aleatoric (predictive uncertainty given the model parameter, i.e., noise in the output) and the epistemic uncertainty (uncertainty about model parameters). We can capture both of these into our interpretable model, since it is fitted to match the reference posterior predictive distribution. The former is captured by conditioning the interpretable model on a posterior draw from the reference model, while the latter is captured by fitting the interpretable model on multiple posterior draws. Details will be given later in Section \ref{Sec:method}. Here, we demonstrate with an example that the proposed method can provide useful information about model uncertainty.
	
	\subsubsection{Practical Example}
	
	\begin{figure}[t]
		\centering
		\includegraphics[width=0.8\linewidth]{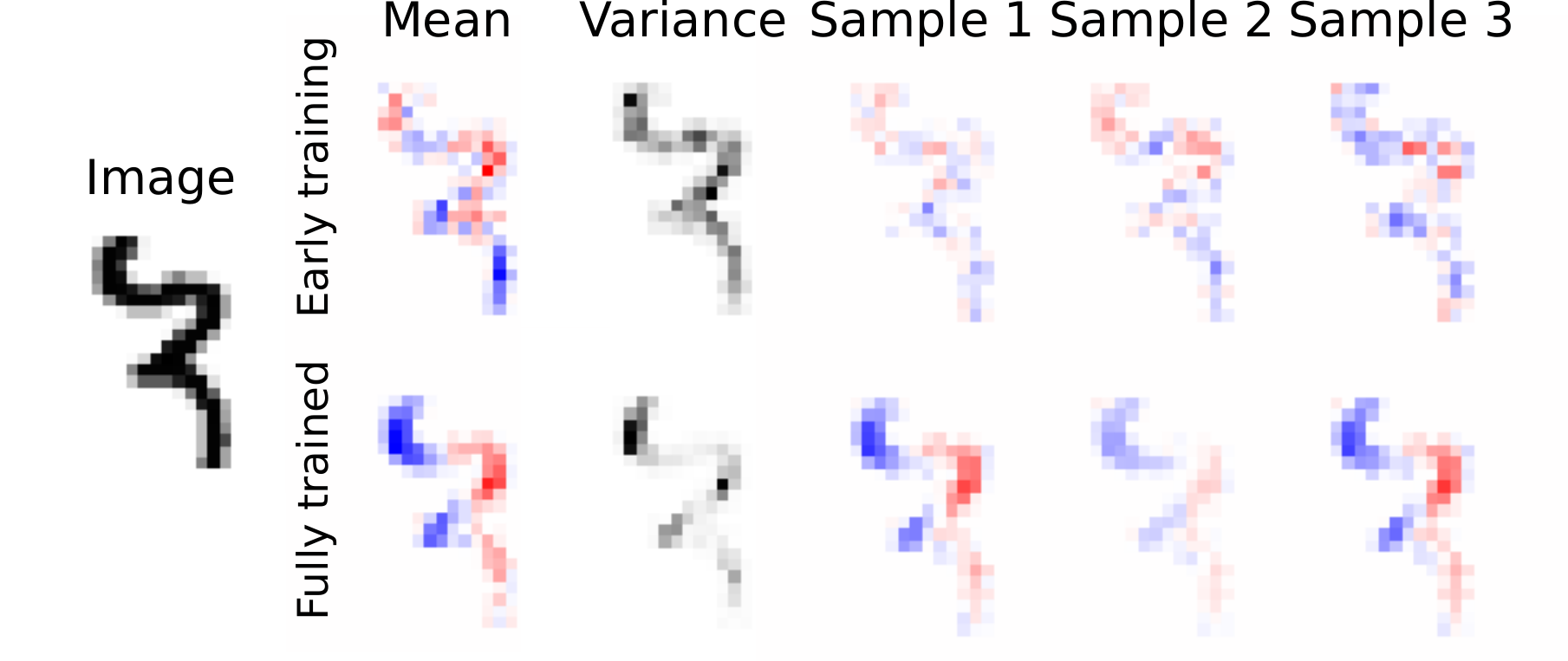}
		\caption{Mean explanation, explanation variance, and three sample explanations for a convolutional neural network 3-vs-8 MNIST-digit classifier early in the training and fully trained. Colored pixels show linear explanation model weights, with red being positive for 3 and blue for 8.}\label{fig:uncertainty_example}
	\end{figure}
	
	We demonstrate uncertainty interpretation in locally explaining a prediction of a Bayesian deep convolutional neural network in the MNIST dataset of images of digits \cite{lecun1998gradient}. The reference model is classifying between digits $3$ and $8$. We use the Bernoulli dropout method \cite{gal2016dropout,gal2016Bayesian}, with a dropout probability of 0.2 and 20 Monte Carlo samples at test time, to approximate Bayesian neural network inference (the posterior predictive distribution). Logistic regression is used as the interpretable model family\footnote{The optimization of the interpretable model follows the general framework explained in Section 3, with logistic regression used as the interpretable model family instead of CART. No penalty for complexity was used here, since the logistic regression model weights are easy to visualize as pseudo-colored pixels}.
	
	Since we are classifying images, we can conveniently visualize the explanation model. Figure~\ref{fig:uncertainty_example} shows visually the logistic regression weights for a digit, comparing the reference model in an early training phase (upper row) and fully trained (lower row). The mean explanations show that the fully trained model has spatially smooth contributions to the class probability, while the model in early training is noisy. Moreover, being able to look at the explanations of individual posterior predictive samples illustrates the epistemic uncertainty. For example, the reference model in early training has not yet been able to confidently assign the upper loop to either indicate a 3 or an 8 (samples 1 and 2 have reddish loop, while sample 3 has bluish). Indeed, the variance plot shows that the model variance spreads evenly over the digit. On the other hand, the fully trained model has little uncertainty about which parts of the digit indicate a 3 or an 8, with most model uncertainty being about the magnitude of the contributions.

	\section{Method: Interpretability Utility for 
		Bayesian Predictive Models}\label{Sec:method}
	
	Here we first explain the procedure to obtain interpretability utility for regression tasks. The case of classification models is similar and is explained in Section \ref{Sec:classification}.
	\subsection{Regression models} 
	Let $\mathcal{D}=\{\left(\bm{x}_n, y_n\right)\}_{n=1}^N$ denote a training set of size $N$, where $\bm{x}_i=\left[x_{i1},\cdots,x_{id}\right]^T$ is a $d$-dimensional feature vector and $y_i \in \mathbb{R}$ is the target variable. Assume that a highly predictive (reference) model $\mathcal{M}$ is fitted to the training data without concerning interpretability constraints. Denote the likelihood of the reference model by  $p\left(y\!\mid\!\bm{x},\bm{\theta},\mathcal{M} \right)$ and the posterior distribution $p\left(\bm{\theta}\!\mid\!\mathcal{D}, \mathcal{M}\right)$. Posterior predictive distribution of the reference model obtains as $\;p(\tilde{y}\!\mid\!\mathcal{D}) = \int_{\theta} p(\tilde{y}\!\mid\!\theta)p(\theta\!\mid\!\mathcal{D})d\theta$. Our goal is to find an interpretable model that best explains the behavior of the reference model locally or globally. 
	We introduce an interpretable model family $\mathcal{T}$ with likelihood $p(y\!\mid\!\bm{x},\bm{\eta},\mathcal{T})$ and posterior $p(\bm{\eta}\!\mid\!\mathcal{D},\mathcal{T})$, belongs to a probabilistic model family with parameters $\bm{\eta}$. The best interpretable model is the one closest to the reference model prediction--wise, and at the same time easily interpretable. To measure the closeness of the predictive behavior of the interpretable model to the reference model, we compute the Kullback-Leibler (KL) divergence between their posterior predictive distribution. Assuming we want to locally interpret the reference model, and following simplifications of \cite{piironen2018projective} for computing the KL divergence of posterior predictive distributions, the best interpretable model can be found by optimizing the following utility function:
	\begin{equation}
	\hat{\bm{\eta}} = \arg \min_{\bm{\eta}} \int \pi_{\bm{x}}(\bm{z})\mbox{KL}\left[p(\tilde{y}\!\mid\!\bm{z},\bm{\theta},\mathcal{M}) \parallel p(\tilde{y}\!\mid\!\bm{z},\bm{\eta},\mathcal{T})\right]d\bm{z} + \Omega(\bm{\eta})
	\label{eq1}
	\end{equation}
	where KL denotes the KL divergence, $\Omega$ is the penalty function for the complexity of the interpretable model, and $\pi_{\bm{x}}(\bm{z})$ is a probability distribution defining the local neighborhood around $\bm{x}$, data point the prediction of which is to be explained. Minimization of the KL divergence verifies that the interpretable model has similar predictive performance to the reference model while the complexity penalty cares for the interpretability of the model.
	
	We compute the expectation in Eq. \ref{eq1} with Monte Carlo approximation by drawing $\{\bm{z}_s\}_{s=1}^S$ samples from $\pi_{\bm{x}}(\bm{z})$:
	\begin{equation}
	\hat{\bm{\eta}}^{(l)} = \arg \min_{\bm{\eta}} \frac{1}{S}\sum_{s=1}^S \mbox{KL}\left[p(\tilde{y}_s\!\mid\!\bm{z}_s,\bm{\theta}^{(l)},M) \parallel p(\tilde{y}_s\!\mid\!\bm{z}_s,\bm{\eta},\mathcal{T})\right] + \Omega(\bm{\eta}),
	\label{eq2}
	\end{equation}
	for $l=1,\ldots,L$ posterior draws from $p(\bm{\theta}\!\mid\!\mathcal{D},\mathcal{M})$. Eq. \ref{eq2} can be solved by first drawing a sample $\bm{\theta}^{(l)}$ from the posterior of the reference model and then finding a sample $\bm{\eta}^{(l)}$ from the posterior of the interpretable model that minimizes the objective function. It has been shown in \cite{piironen2018projective} that minimization of the KL-divergence in Eq. \ref{eq2} is equivalent to maximizing the expected log-likelihood of the interpretable model over the likelihood obtained by a posterior draw from the reference model:
	\begin{equation}
	\arg \max_{\bm{\eta}} \; \frac{1}{S}\sum_{s=1}^S E_{\tilde{y}_s\mid\bm{z}_s,\bm{\theta}^{(l)}}\left[\log p\left(\tilde{y}_s\!\mid\!\bm{z}_s,\bm{\eta}\right)\right].
	\label{eq3}
	\end{equation}
	Using this equivalent form and by adding the complexity penalty term, the interpretability utility obtains as
	\begin{equation}
	\arg \max_{\bm{\eta}} \frac{1}{S}\sum_{s=1}^S E_{\tilde{y}_s\mid\bm{z}_s,\bm{\theta}^{(l)}}\left[\log p\left(\tilde{y}_s\!\mid\!\bm{z}_s,\bm{\eta}\right)\right] - \Omega\left(\bm{\eta}\right).
	\label{interp_utility}
	\end{equation}
	The complexity penalty term should be chosen to match the resulting model; possible options are the number of leaf nodes for decision trees, number of rules and/or size of the rules for rule list models, number of non-zero weights for linear regression models, etc. Although the proposed approach is general and can be used for any family of interpretable models, in the following, we use CART models with tree size (the number of leaf nodes) as the measure of interpretability.
	With this assumption, similar to the illustrative example in Section \ref{Sec:ref_illust_expl}, the interpretability constraint is defined over the model space; it could also be defined over the parameter space of a particular model, such as tree shape parameters of Bayesian CART models \cite{chipman1998bayesian}. The interpretability prior approach corresponds to fitting a CART model to the training data, i.e. samples drawn from the neighborhood distribution of $\bm{x}$.
	
	A CART model describes $p(y\!\mid\!\bm{z},\bm{\eta})$ with two main components $\bm{\eta}=\left(T, \bm{\phi}\right)$: a binary tree $T$ with $b$ terminal nodes and a parameter vector $\bm{\phi} = (\phi_1, \phi_2, \cdots, \phi_b)$ that associates the parameter value $\phi_i$ with the $i$th terminal node. If $\bm{z}$ lies in the region corresponding to the $i$th terminal node, then $y\!\mid\!\bm{z},\bm{\eta}$ has distribution $f(y\!\mid\!\phi_i)$, where $f$ denotes a parametric probability distribution with parameter $\phi_i$. For CART models, it is typically assumed that, conditionally on $\bm{\eta}$, values $y$ within a terminal node are independently and identically distributed, and $y$ values across terminal nodes are independent. In this case, the corresponding likelihood of the interpretable model for the $l$th draw from the posterior of $\bm{\theta}$ has the form
	\begin{equation}
	p\left(\bm{y}\!\mid\!\bm{Z},\bm{\eta}^{(l)}\right)=\prod_{i=1}^b f\left(\bm{y}_i\!\mid\!\phi_i^{(l)}\right)=\prod_{i=1}^b\prod_{j=1}^{n_i}f\left(y_{ij}\!\mid\!\phi_i^{(l)}\right),
	\label{cart_ll}
	\end{equation}
	where $\bm{y}_i \equiv (y_{i1},\cdots,y_{in_i})$ denotes the set of the $n_i$ observations assigned to the partition generated by the $i$th terminal node with parameter $\phi_i^{(l)}$, and $\bm{Z}$ is the matrix of all the $\bm{z}_s$. For regression problems, assuming a mean-shift normal model for each terminal node $i$,\footnote{In the mean-variance shift model, each terminal node has its own $\sigma_i^2$ variable and the number of parameters is $2\times b$.} 
	the likelihood of the interpretable model is defined as
	\begin{equation}
	f\left(\bm{y}\mid \bm{\phi}^{(l)}\right) = \prod_{i=1}^{b}\prod_{j=1}^{n_i} N\left(y_{ij}\!\mid\!\mu_i^{(l)}, {\sigma^2}^{(l)}\right),
	\label{leaf_mdl}
	\end{equation}
	where $\bm{\phi}^{(l)}=(\bm{\mu}^{(l)}=\{\mu_i^{(l)}\}_{i=1}^b, {\sigma^2}^{(l)})$. With this formulation, the task of finding an interpretable proxy to the reference model $M$ is reformed to find a tree structure $T$ with parameters $\bm{\phi}^{(l)}$ such that its predictive performance is as close as possible to $M$, while being as interpretable as possible. Interpretability is measured by the complexity term $\Omega$.
	
	The log-likelihood of the tree with the $S$ samples drawn from the neighborhood of $\bm{x}$ is
	\begin{equation}
	\mathcal{L} = -\frac{S}{2}\log(2\pi\sigma^2)-\frac{1}{2\sigma^2}\sum_{i=1}^b \sum_{j=1}^{n_i} \left(y_{ij}-\mu_i\right)^2.
	\label{exp_mdl_ll}
	\end{equation}
	Projecting this into Eq. \ref{interp_utility}, the interpretability utility has the following form:
	\begin{equation}
	\begin{aligned}
	\MoveEqLeft \arg \max_{\bm{\eta}} \; -\frac{1}{2}\log(2\pi\sigma^2)-\frac{1}{2S\sigma^2}\sum_{i=1}^b\sum_{j=1}^{n_i} E_{y_{ij}\mid \bm{\theta}^{(l)}}\left[\left(y_{ij}-\mu_i\right)^2\right] - \Omega(T) \\
	\MoveEqLeft \propto \arg \max_{\bm{\eta}} \;-\frac{1}{2}\log(2\pi\sigma^2)-\frac{1}{2S\sigma^2}\sum_{i=1}^b\sum_{j=1}^{n_i}\left[\sigma_{ij}^2 + \left(\Bar{y}_{ij}-\mu_i\right)^2\right] - \Omega(T),
	\end{aligned}
	\label{cart_interp_utility}
	\end{equation}
	where $\Bar{y}_{ij}$ and $\sigma_{ij}^2$ are respectively the mean and variance of the reference model for the $j$th sample in the $i$th terminal node. $\Omega(T)$ is a function of the interpretability of the CART model. Here we set it to $\alpha b$ using $\alpha$ as a regularization parameter. The pseudocode of the proposed approach is shown in Algorithm \ref{alg1}.
	\begin{figure}[ht]
		\centering
		\begin{minipage}{\linewidth}
			\begingroup
			\removelatexerror
			\RestyleAlgo{boxruled}
			\begin{algorithm}[H]
				\SetAlgoLined
				\KwInput{training data $\mathcal{D}=\{\left(\bm{x}_n,y_n\right)\}_{n=1}^N$, a test sample $\bm{x}_{test}$ to be explained}
				\KwOutput{a decision tree explaining the prediction for the test sample $\bm{x}_{test}$}
				\vspace{0.5cm}
				\tcc{REFERENCE MODEL CONSTRUCTION}
				fit the Bayesian predictive model to $\mathcal{D}$ without concerning interpretability constraints\;
				draw $\{\bm{z}_s\}_{s=1}^S$ from the neighborhood of $\bm{x}_{test}$ defined by $\pi_{\bm{x}}$\;
				\For{each draw $\bm{z}_s$}{
					get the mean $\Bar{y}_s$ and variance $\sigma^2_s$ of the Bayesian predictive distribution\;
				}
				\tcc{INTERPRETABILITY OPTIMIZATION}
				fit a CART model to $\{\left(\bm{z}_s,\Bar{y}_s\right)\}_{s=1}^S$ by optimizing Eq. \ref{cart_interp_utility} 
				\caption{Decision--theoretic approach for local interpretability in the Bayesian framework}
				\label{alg1}
			\end{algorithm}
			\endgroup
		\end{minipage}
	\end{figure}
	
	When fitting a global interpretable model, instead of drawing samples from $\pi_{\bm{x}}$, we use training inputs $\{\bm{x}_n\}_{n=1}^N$ with their corresponding output computed by the reference model $\{y_n^{ref}\}_{n=1}^N$ as the target value.
	
	The next subsection explains how to solve Eq. \ref{cart_interp_utility} for CART models.
	
	\subsection{Optimization Approach}\label{opt_prc}
	We optimize Eq. \ref{cart_interp_utility} by using the backward fitting idea which involves first growing a large tree and then pruning it back to obtain a smaller tree with better generalization. For this goal, we use the formulation of maximum likelihood regression tree (MLRT) \cite{su2004maximum}.
	\subsubsection{Growing a large tree}
	Given the training data\footnote{Here, for local interpretation, training data refers to the $S$ samples (with their corresponding predictions made by the reference model) taken from the neighborhood distribution to fit the explainable model.}, MLRT automatically decides on the splitting variable $x_j$ and split point (a.k.a. pivot) $c$ using a greedy search algorithm that aims to maximize the log-likelihood of the tree by splitting the data in the current node into two parts: the left child node satisfying $x_j \leq c$ and the right child node satisfying $x_j > c$. The procedure of growing the tree is as follows:
	\begin{itemize}
		\item[1.] For each node $i$, determine the maximum likelihood estimate of its mean parameter $\mu_i$ given observations associated with the node, and then compute the variance parameter of the tree given $\{\mu_i\}_{i=1}^b$: 
		\begin{equation*}
			\begin{split}
				\Hat{\mu}_i = \frac{1}{n_i}\sum_{j=1}^{n_i}\Bar{y}_{ij} \\
				\Hat{\sigma}^2 = \frac{\sum_{i=1}^b\sum_{j=1}^{n_i} \left[\sigma_{ij}^2+\left(\Bar{y}_{ij} - \hat{\mu}_i\right)^2\right]}{S}.
			\end{split}
			\label{eq8}
		\end{equation*}
		The log-likelihood score of the node is then computed, up to a constant, by $\mathcal{L}_i \propto - n_i\log (\hat{\sigma}^2)$. 
		\item[2.] For each variable $x_j$, determine the amount of increase in the log-likelihood of the node $i$ caused by a split $r$ as
		\begin{equation*}
			\Delta_{(r,x_j,i)}= \mathcal{L}_{i_R} + \mathcal{L}_{i_L} - \mathcal{L}_{i},
			\label{eq10}
		\end{equation*}
		where $\mathcal{L}_{i_R}$ and $\mathcal{L}_{i_L}$ are the log-likelihood scores of the right and left child nodes of the parent node $i$ generated by the split $r$ on the variable $x_j$, respectively.
		\item[3.] For each variable $x_j$, select the best split $r_j^*$ with largest increase to the log-likelihood. 
		\item[4.] Among the best splits, the one that causes the global maximum increase in the log-likelihood score will be selected as the global best split, $r^*$, for the current node, i.e. $r^* = \max_{r_j^*,\;j=1,\cdots,d} \Delta_{\left(r_j^*,x_j,i\right)}$.
		\item[5.] Iterate steps 1 to 4 until reaching the stopping criteria.
	\end{itemize}
	In our implementation, we used the minimum size of a terminal node (the number of samples lie in the region generated by the terminal node) as the stopping condition.
	\subsubsection{Pruning}
	We adopt the cost-complexity pruning using the following cost function:
	\begin{equation}
	C_{\alpha}\left(T\right) = \log(\hat{\sigma}^2) + \alpha b.
	\label{eq11}
	\end{equation}
	Pruning is done iteratively; in each iteration $i$, the internal node $h$ that minimizes $\alpha = \frac{\left(C(h) - C(T_i)\right)}{\left(\mid\mbox{leaves}\left(T_h\right)\mid - 1\right)}$ is selected for pruning, where $C(h)$ refers to the cost of the decision tree with $h$ as terminal node, $C(T_i)$ denote the cost of the full decision tree in iteration $i$, and $T_h$ denotes the subtree with $h$ as its root. The output is a sequence of decision trees 
	and a sequence of $\alpha$ values. 
	The best $\alpha$ and its corresponding subtree are selected using $5$-fold cross-validation.
	
	\subsection{Classification models}\label{Sec:classification}
	For classification problems, assuming the CART models as the interpretable model family, the form of the interpretability utility is the same as Equation \ref{interp_utility} except that the likelihood of the interpretable model follows a multinomial distribution with the following log-likelihood:
	\begin{equation}
	\mathcal{L} = \sum_{i=1}^{b}\sum_{j=1}^{n_i}\sum_{k=1}^{K}I\left(y_{ij}\in C_k\right)\log p_{ik} \;\;\; \mbox{s.t.}\;\;
	p_{ik}\geq 0, \;\; \sum_{k=1}^{K}p_{ik}=1
	\end{equation}
	where $I(y_{jk}\in C_k)$ is the indicator function determining wheter or not the $j$\textsuperscript{th} sample of the $i$\textsuperscript{th} node belongs to the $k$\textsuperscript{th} category assuming that there are in total $K$ categories. The $p_{ik}$ denote the probability of the occurrence of the $k$\textsuperscript{th} category in the $i$\textsuperscript{th} terminal node and the set of parameters are $\bm{\phi}=\{\bm{p}_i=\left(p_{i1},\ldots,p_{ik}\right)\}_{i=1}^b$. Therefore, the final form of the interpretability utility for Bayesian classification models is
	\begin{equation}
	\arg \max_{\bm{\eta}} \; \frac{1}{S}\sum_{i=1}^b\sum_{k=1}^{K} n_k\log p_{ik} + \Omega(T) \;\; \mbox{s.t.} \;\; p_{ik}\geq 0, \;\; \sum_{k=1}^{K}p_{ik}=1
	\end{equation}
	where $\bm{\eta} = \left(T,\bm{\phi}\right)$ and $n_k = \sum_{j=1}^{n_i}I\left(y_{jk}\in C_k\right)$. The optimization approach is again similar to the process explained in Section \ref{opt_prc} with the difference that the maximum likelihood estimate of the parameters of each node $i$ obtains as $\hat{p}_{ik}=\frac{n_k}{n_i}$. Finally, the log-likelihood score of each node $i$ is determined by $\mathcal{L}_i = \sum_{k=1}^{K}n_k\log\hat{p}_{ik}$.
	
	\subsection{Connection With Local Interpretable Model-agnostic Explanation (LIME)}
	LIME \cite{ribeiro2016should} is a prediction-level interpretation approach that fits a sparse linear model to the black-box model's prediction via drawing samples from the neighborhood of the data point to be explained. Our proposed approach extends LIME to KL divergence based interpretation of Bayesian predictive models (although it can also be used for non-Bayesian probabilistic models as well). This is achieved by combining the idea of LIME with the idea of projection predictive variable selection \cite{piironen2018projective}. The approach is able to handle different types of predictions (continuous valued, class labels, counts, censored and truncated data, etc.) and interpretations (model-level or prediction-level) as long as we can compute KL divergence between the predictive distributions of the original model and the explanation model. For a more detailed explanation of the connection, check the preliminary work of \cite{peltola2018local}.
	
	\section{Experiments}\label{Sec.exp}
	
	We demonstrate the efficacy of the proposed approach through experiments on several real-world data sets. Subsection \ref{subsec:global_exp} discusses the experiments related to global intepretation. We first investigate the effect of reference models with different predictive powers on the performance of the final interpretable model. Secondly, we compare our approach with the interpretability prior alternative, of fitting directly an interpretable model to the data, in terms of their capability to trade off between accuracy and interpretability. We also compare the performance of our approach with non-Bayesian counterparts introduced in Section \ref{Sec:intro}. Further, we investigate the stability of our approach and the interpretability prior approach. Section \ref{subsec:local_exp} examines local interpretation, where we compare our approach with LIME. Our codes and data are available online at \href{https://github.com/homayunafra/Decision_Theoretic_Approach_for_Interpretability_in_Bayesian_Framework}{github.com/homayunafra/Decision\_Theoretic\_Approach\_for\_Interpretability\_in\_Bayesian\_Framework}.
	\begin{figure}[t!]
		\centering
		\begin{tabular}[b]{c}
			\includegraphics[scale=.25]{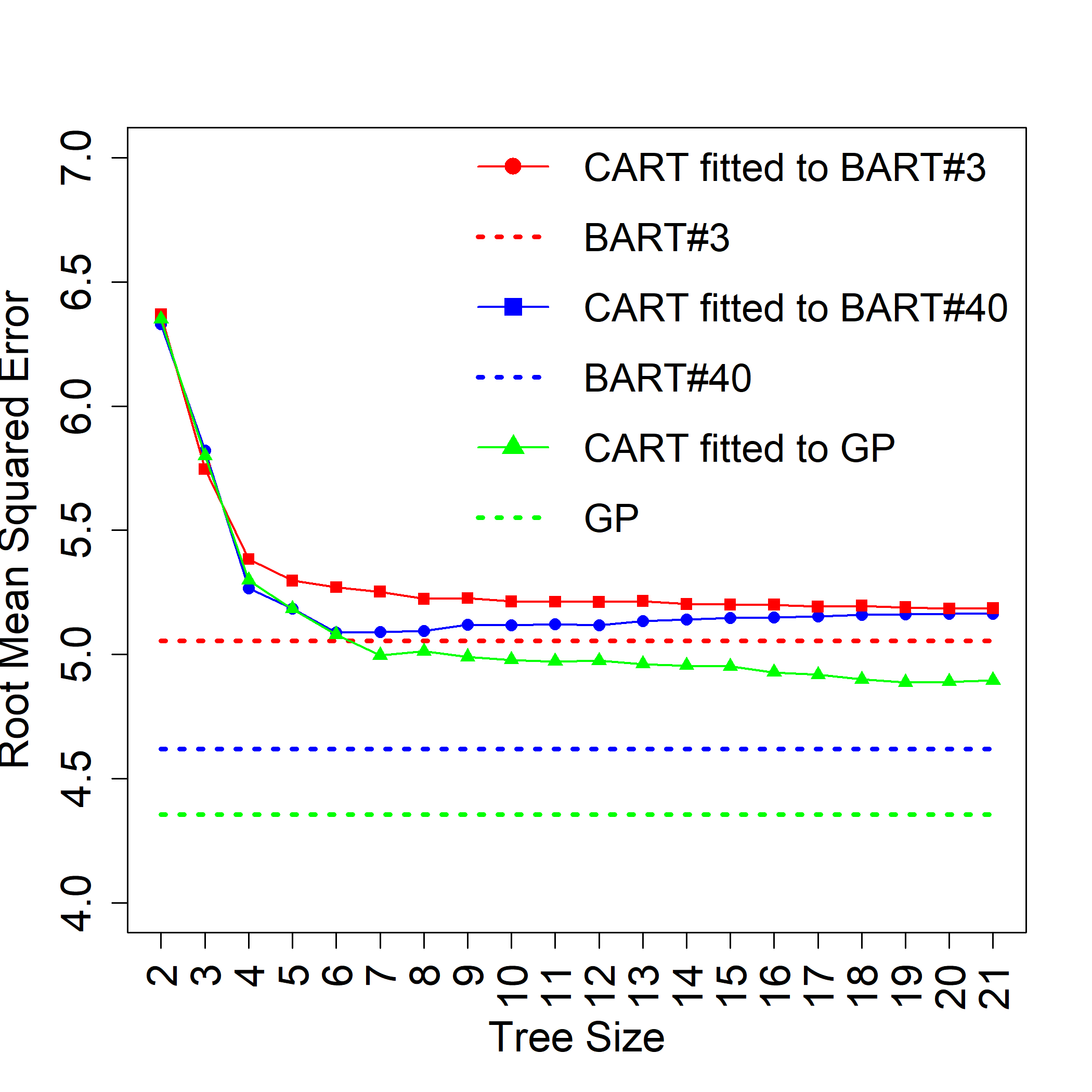} \\
			\small (a) Body fat
		\end{tabular} \hspace{-8pt}
		\begin{tabular}[b]{c}
			\includegraphics[scale=.25]{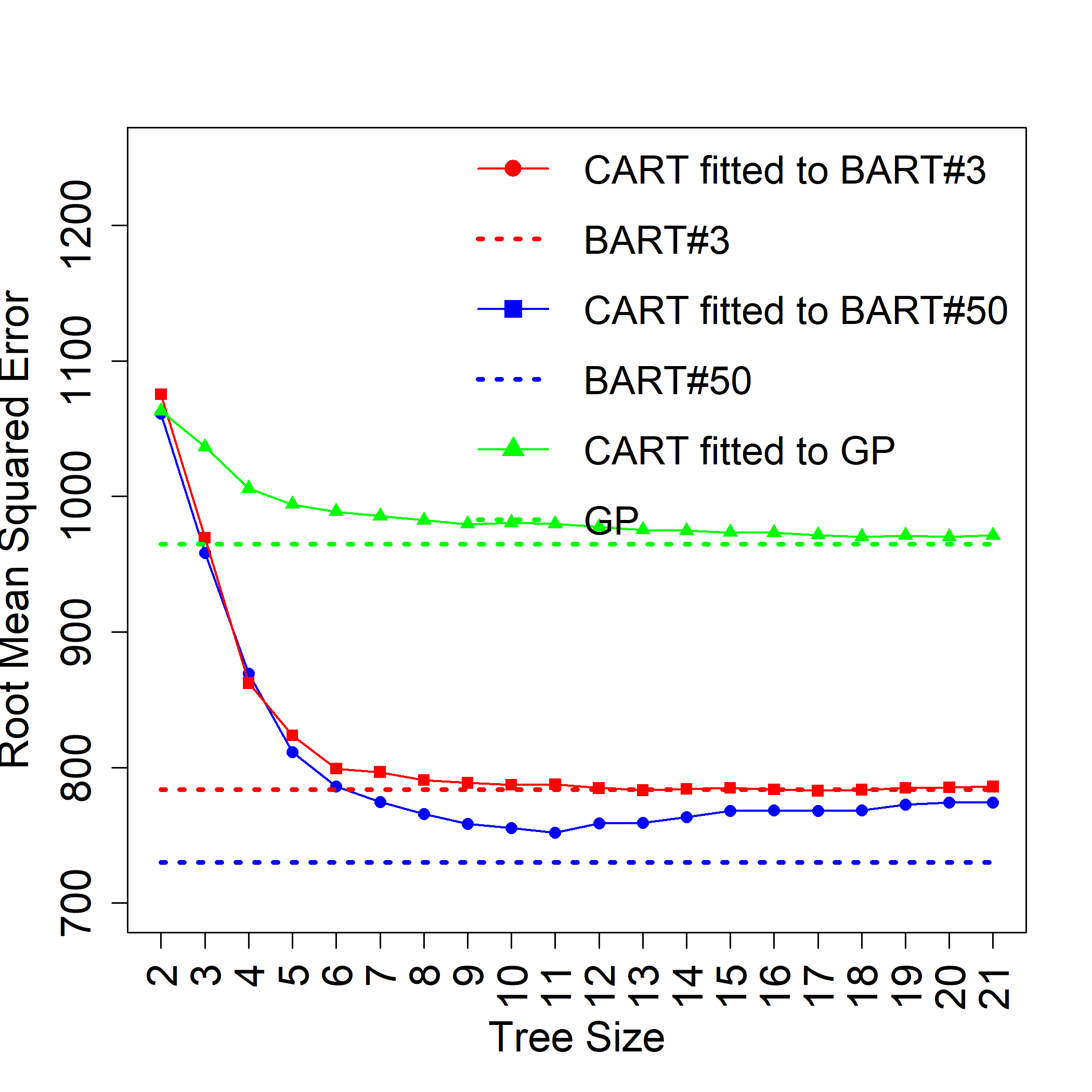} \\
			\small (b) Baseball
		\end{tabular}\hspace{-8pt}
		\begin{tabular}[b]{c}
			\includegraphics[scale=.25]{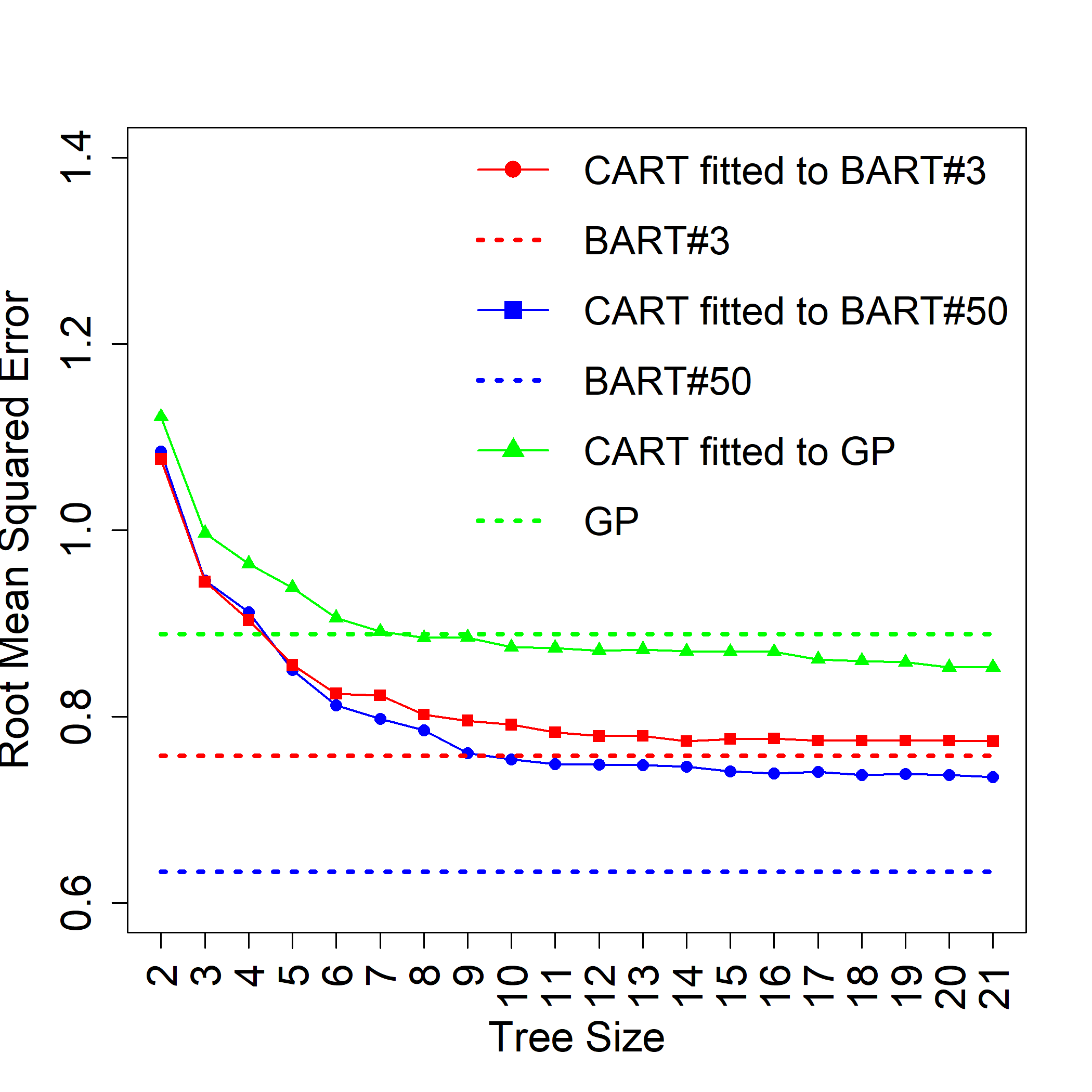} \\
			\small (c) Auto risk
		\end{tabular}
		\caption{Effect of reference models with different predictive performance on the performance of the interpretable models fitted to them. More accurate reference models result in interpretable models with higher predictive performance. The values of ``ntree'' used for the BART models are shown by ``\#''.}
		\label{fig:ref_effect}
	\end{figure}
	\subsection{Global Interpretation}\label{subsec:global_exp}
	
	\subsubsection{Data}\label{subsec:data}
	
	In our experiments, we use the following data sets: body fat \cite{johnson1996fitting}, baseball players \cite{hoaglin1995critical}, auto risk \cite{kibler1989instance}, bike rental \cite{fanaee2014event}, auto mpg \cite{quinlan1993combining}, red wine quality \cite{cortez2009modeling}, and Boston housing \cite{harrison1978hedonic}.
	Each data set is divided into training and test set containing $75\%$ and $25\%$ of samples, respectively. 
	
	\subsubsection{Effect of Reference Model}
	
	The purpose of this test is to evaluate how the predictive power of the reference model affects the performance of the interpretable model when it is used to globally explain the reference model. Three data sets are adopted for this test: body fat, baseball players, and auto risk. Furthermore, three reference models with different predictive powers are adopted: two BART models, and a Gaussian process (GP).
	
	For the BART models, we used the BART package in R with two different values for the ``ntree'' (number of trees) parameter. For one model, ``ntree'' is set to the value that gives the highest predictive performance on the validation set (blue dotted line in Figure \ref{fig:ref_effect}), while for another one, this parameter is set to $3$, a low value, which gives poor predictive performance (red-dotted line in Figure \ref{fig:ref_effect}). The rest of the parameters are set to their default values except ``nskip" and ``ndpost'', which are set to $2000$ and $4000$, respectively. For the BART models, mean of the predictions of the posterior draws is used as their output. For the GP (green-dotted line in Figure \ref{fig:ref_effect}), ``Matern52'' is used as the kernel with variance and length scales obtained by cross-validation over a small grid of values\footnote{This may not be the best setting for the GP. We did not attempt to optimize that since our objective is not to compare the performance of GP with BART, but instead to compare the performance of the interpretable models fitted to them.}.
	\begin{figure}[t]
		\centering
		\begin{tabular}[b]{c}
			\includegraphics[scale=.25]{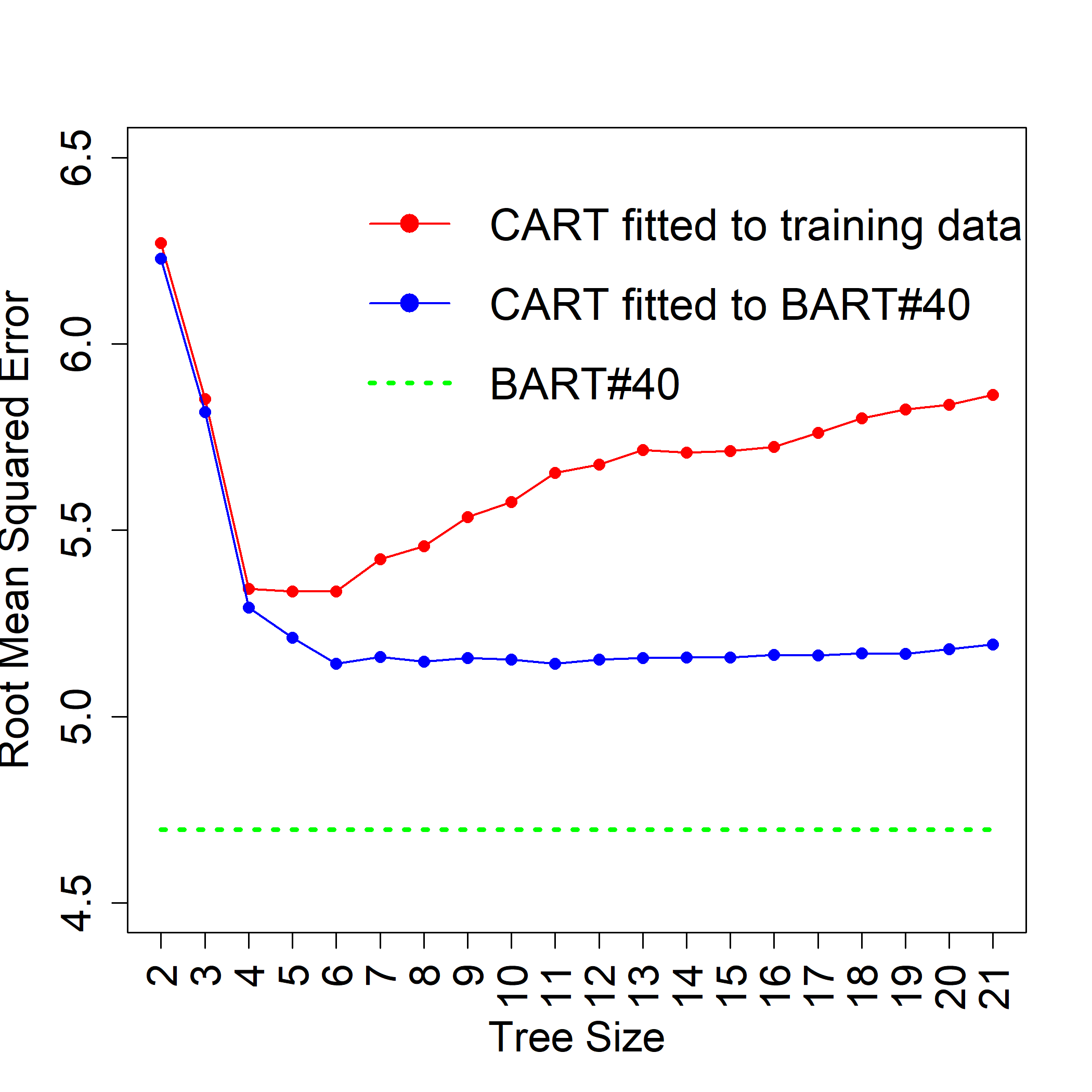} \\
			\small (a) Body fat
		\end{tabular} \hspace{-8pt}
		\begin{tabular}[b]{c}
			\includegraphics[scale=.25]{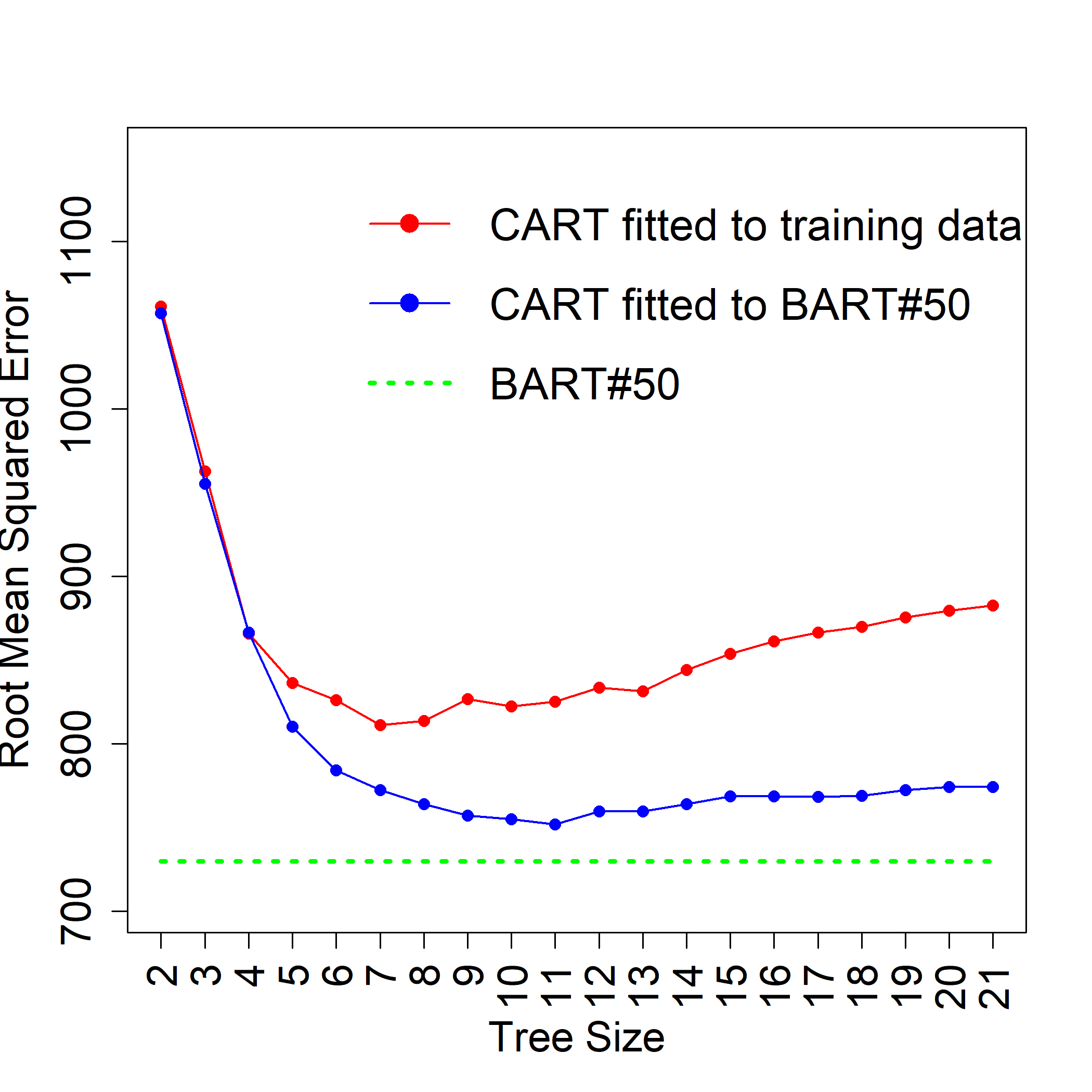} \\
			\small (b) Baseball
		\end{tabular}\hspace{-8pt}
		\begin{tabular}[b]{c}
			\includegraphics[scale=.25]{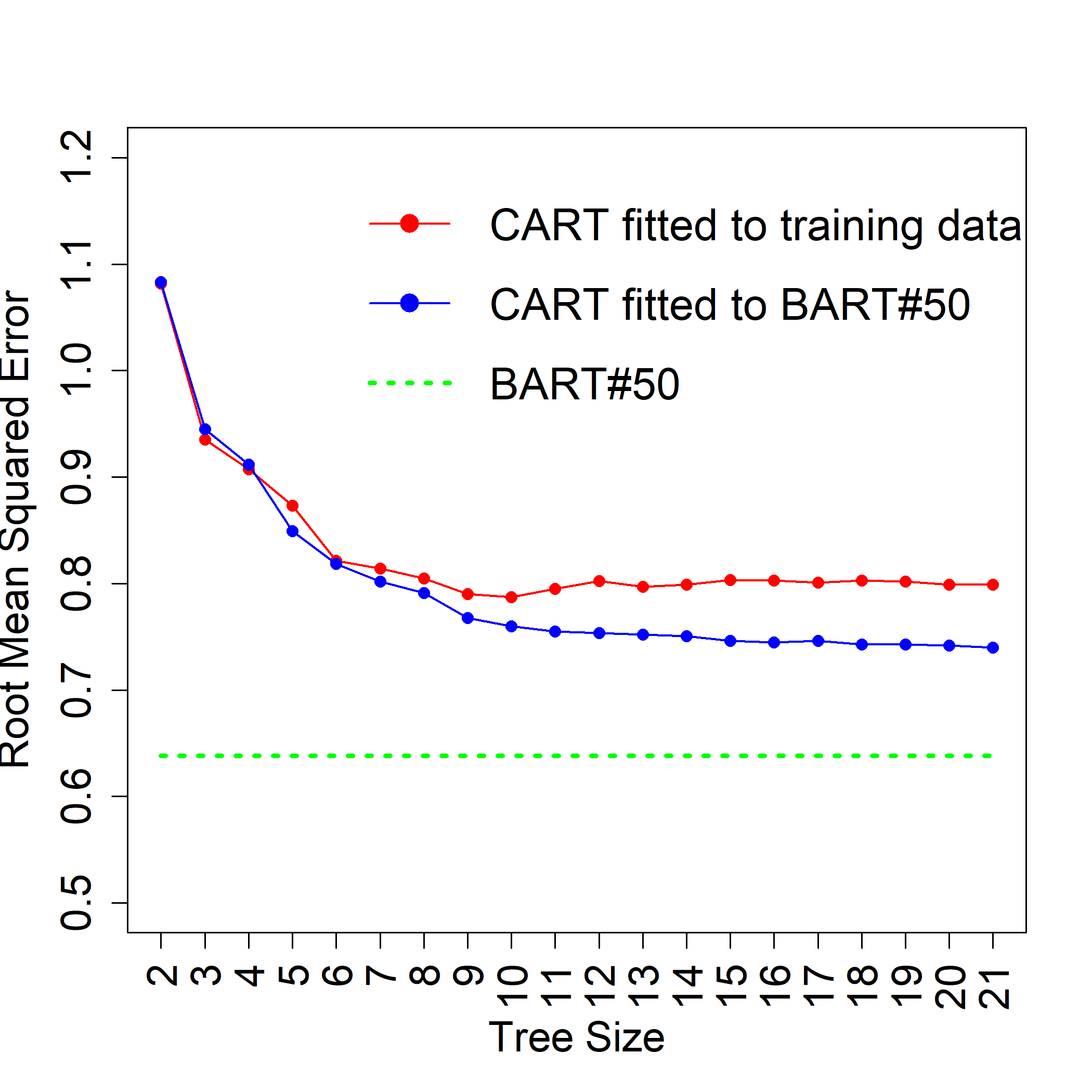} \\
			\small (c) Auto risk
		\end{tabular}\\
		\begin{tabular}[b]{c}
			\includegraphics[scale=.25]{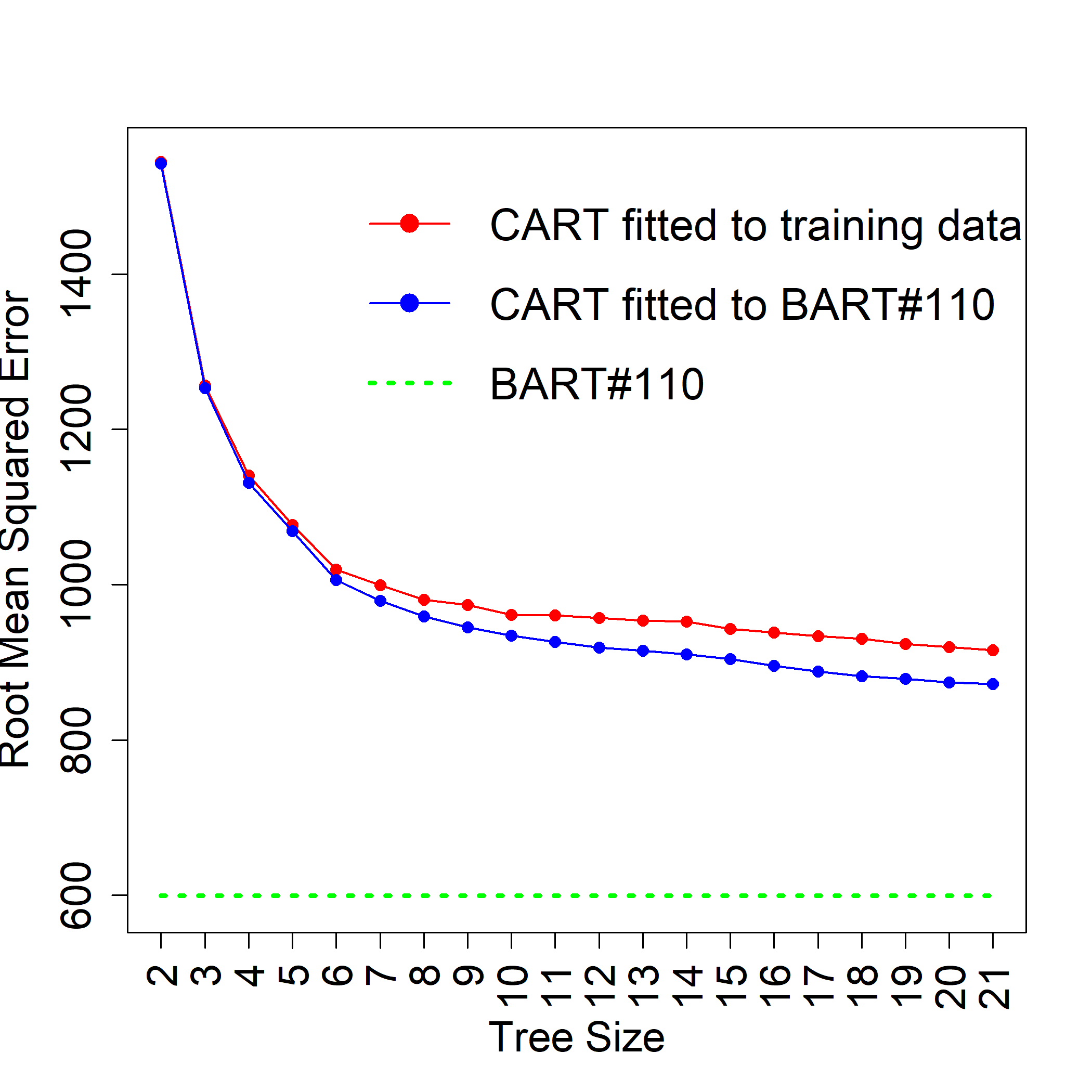} \\
			\small (d) Bike
		\end{tabular} \hspace{-8pt}
		\begin{tabular}[b]{c}
			\includegraphics[scale=.25]{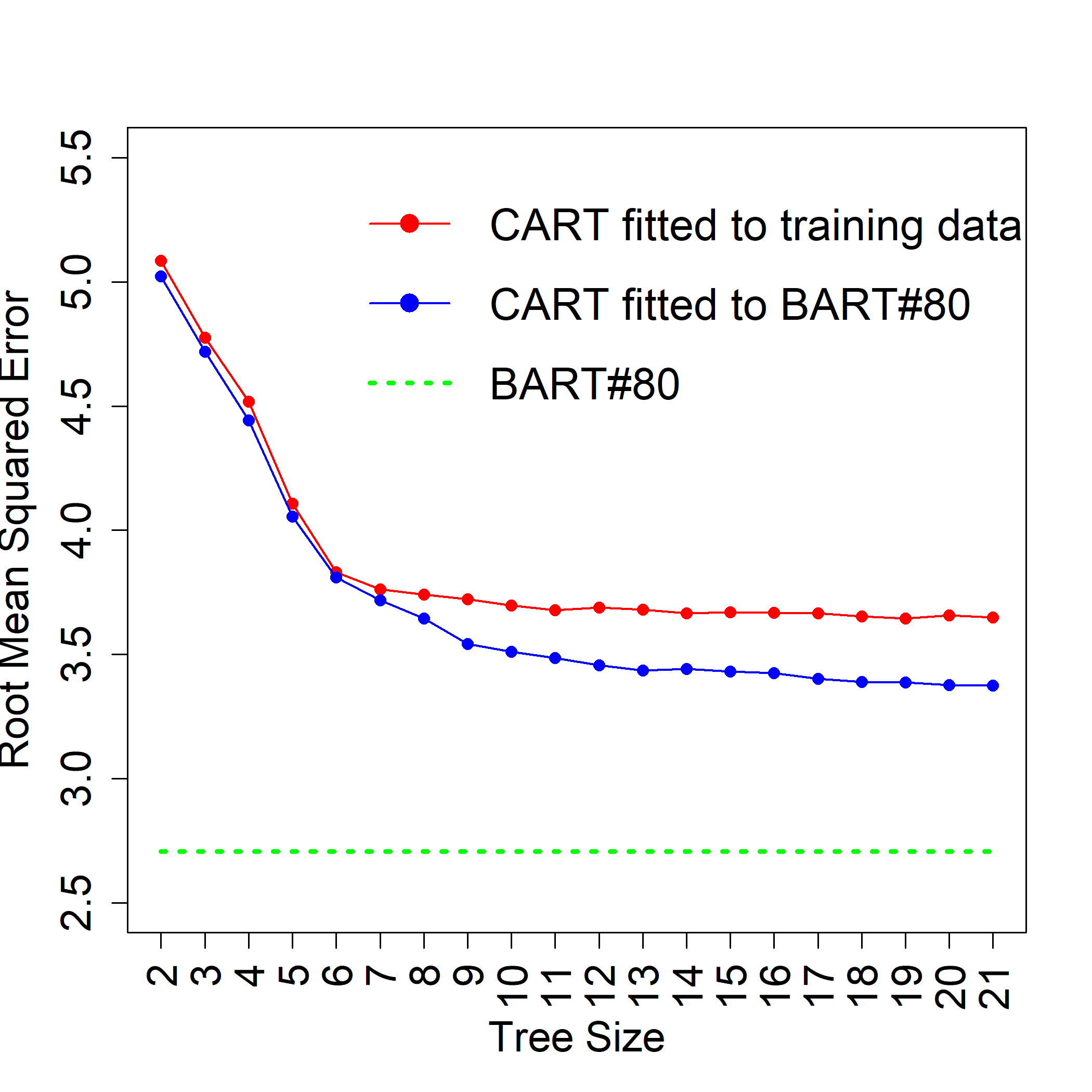} \\
			\small (e) Auto mpg
		\end{tabular}\hspace{-8pt}
		\begin{tabular}[b]{c}
			\includegraphics[scale=.25]{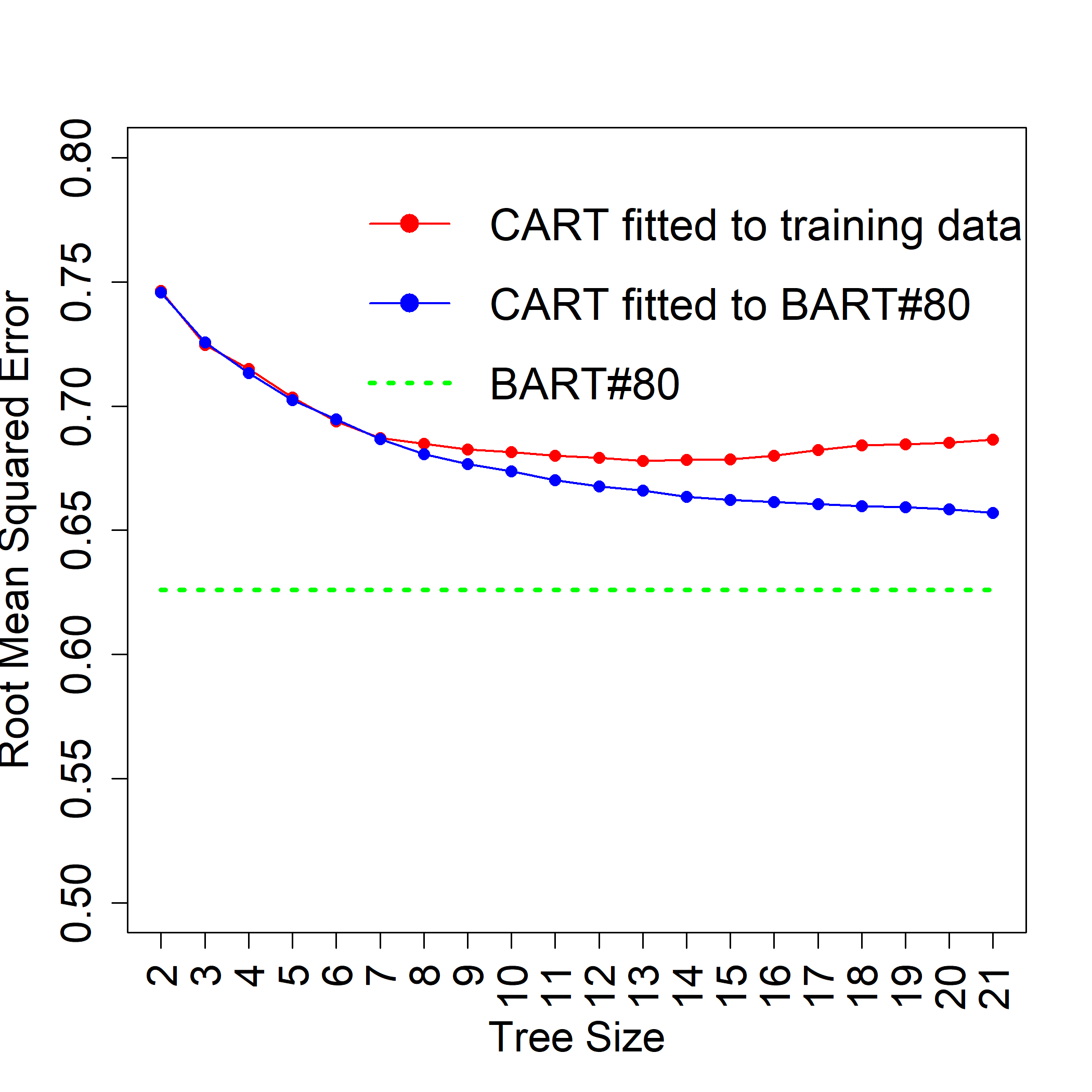} \\
			\small (f) Wine quality
		\end{tabular}
		\caption{Comparison of interpretability prior (red) and interpretability utility (blue) approach in trading off between accuracy and interpretability when using CART as explainable models and BART as reference model. The values of ``ntree'' used for the BART models are shown by ``\#''.}
		\label{perf_comp}
	\end{figure}
	CART models are used as the interpretable model family. The size of the tree, i.e., the total number of leaf nodes, is used as the measure of interpretability \cite{bastani2018interpreting, hara2016making}. 
	
	Figure \ref{fig:ref_effect} demonstrates the results, which are averaged over $50$ runs. The difference in the predictive performance of the interpretable models fitted to different reference models suggests that using more accurate reference models (BART in Baseball and Auto risk data sets, and GP in Body fat data set) can generate more accurate interpretable models as well. This is expected since by the performance of the interpretable model converges to the performance of the reference model; therefore the interpetable model will be more accurate when fitted to a more accurate reference model. The gap between the predictive performance of the interpretable models and their corresponding reference models is due to the limited predictive capability of the interpretable model. For some tasks, this gap can be made narrower by increasing the complexity of the interpretable model, while for others, a different family of interpretable models may be needed. 
	
	Finally, in Figure \ref{fig:ref_effect}.c, the performance of the interpretable model fitted to the GP reference model is better than the reference model itself, for some complexities. This may be because of different extrapolation behavior of CART and GP. In the high-dimensional space, the test data may be outside the support of the training data; thus, extrapolation behavior matters. Simpler models can make more conservative extrapolations which may be helpful in this case.
	\begin{figure}[t!]
		\centering
		\begin{tabular}[b]{c}
			\includegraphics[scale=.25]{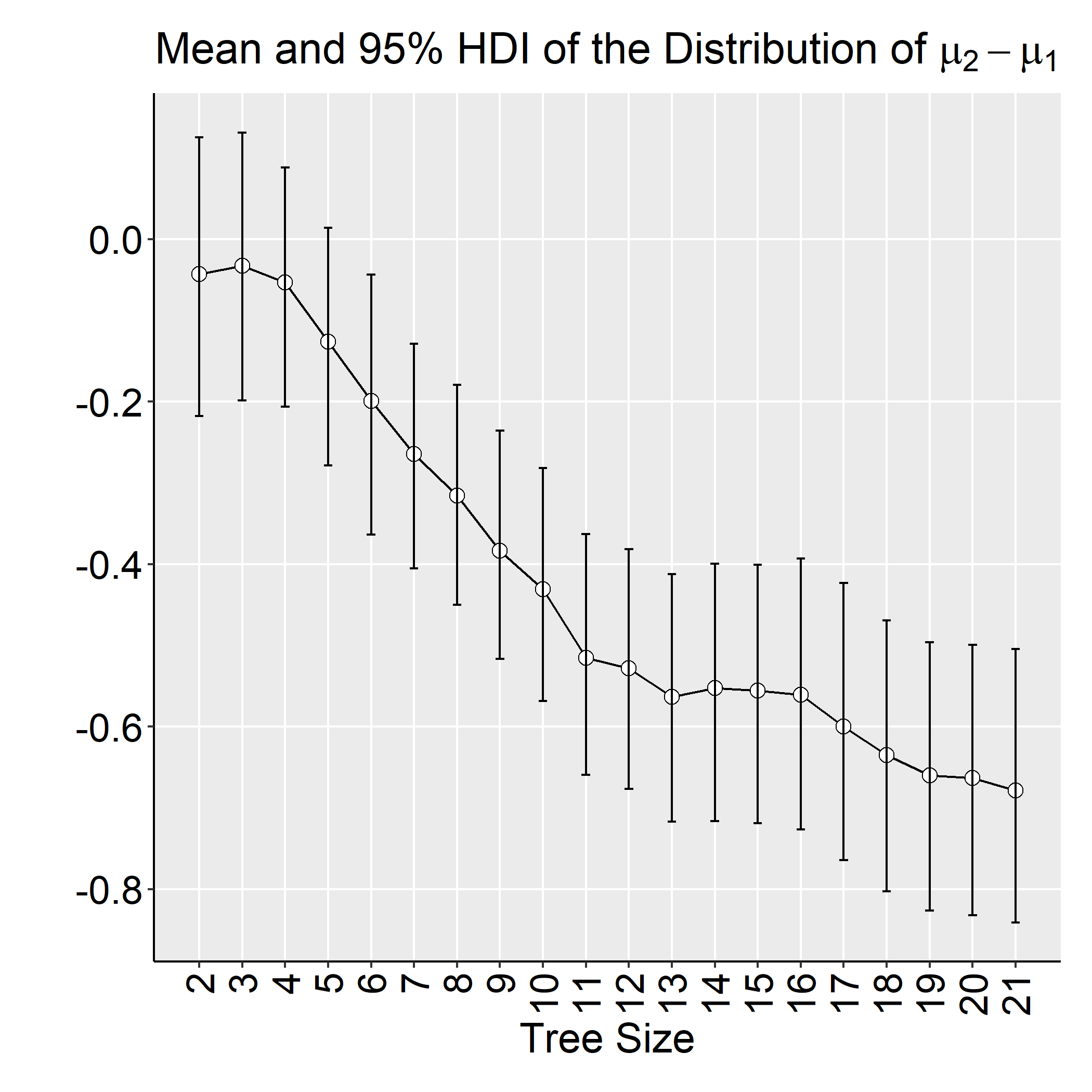} \\
			\small (a) Body fat
		\end{tabular} \hspace{-8pt}
		\begin{tabular}[b]{c}
			\includegraphics[scale=.25]{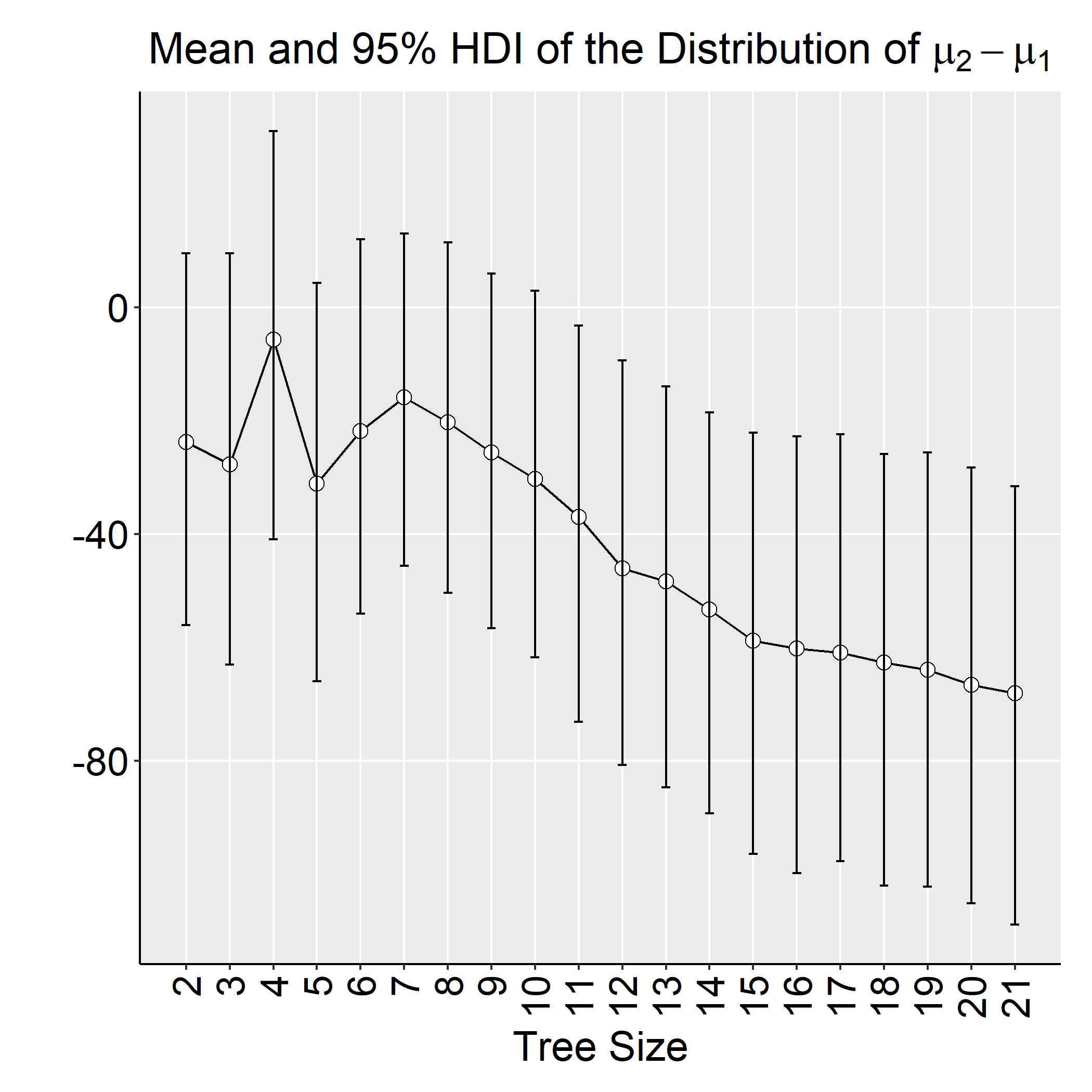} \\
			\small (b) Baseball
		\end{tabular}\hspace{-8pt}
		\begin{tabular}[b]{c}
			\includegraphics[scale=.25]{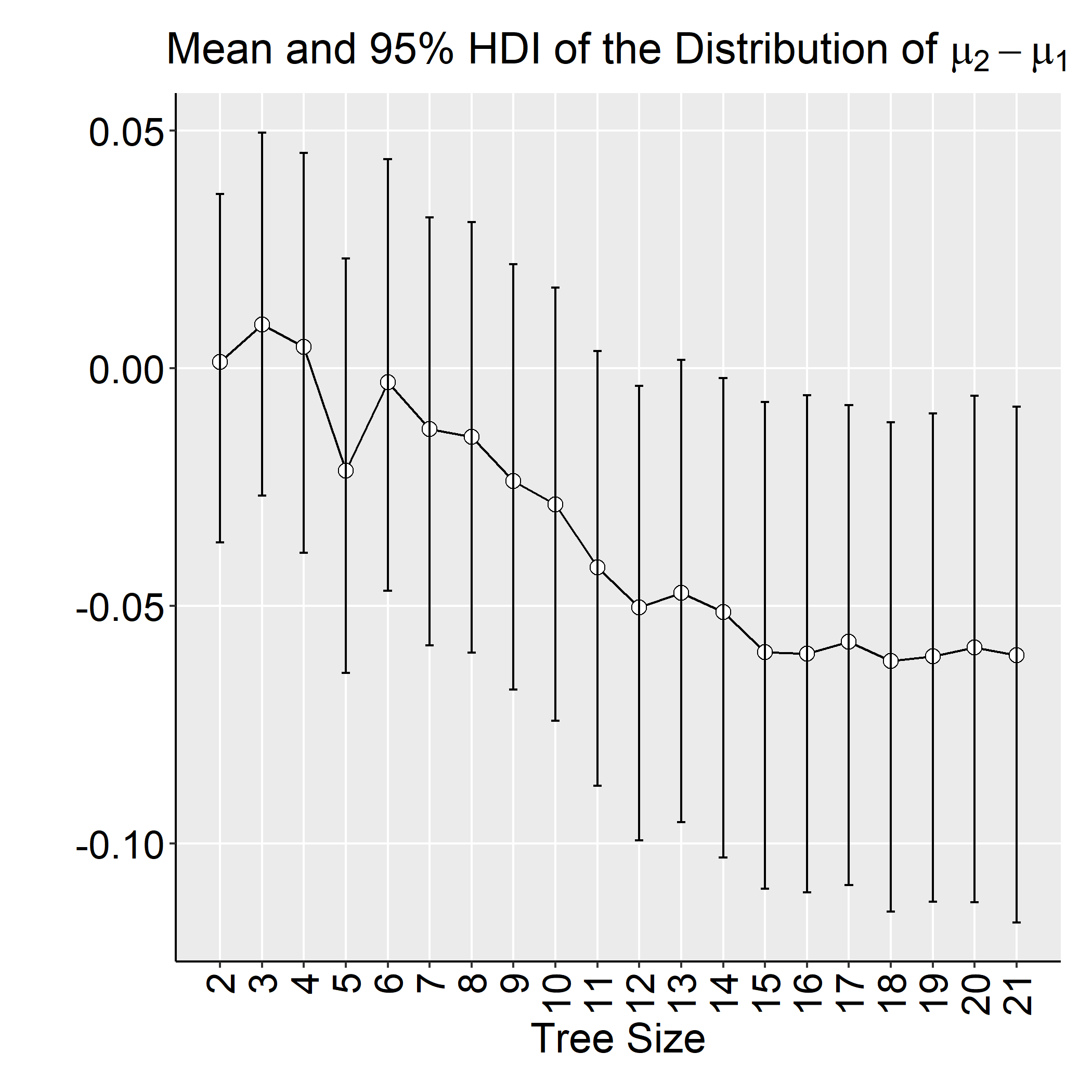} \\
			\small (c) Auto risk
		\end{tabular}\\
		\begin{tabular}[b]{c}
			\includegraphics[scale=.25]{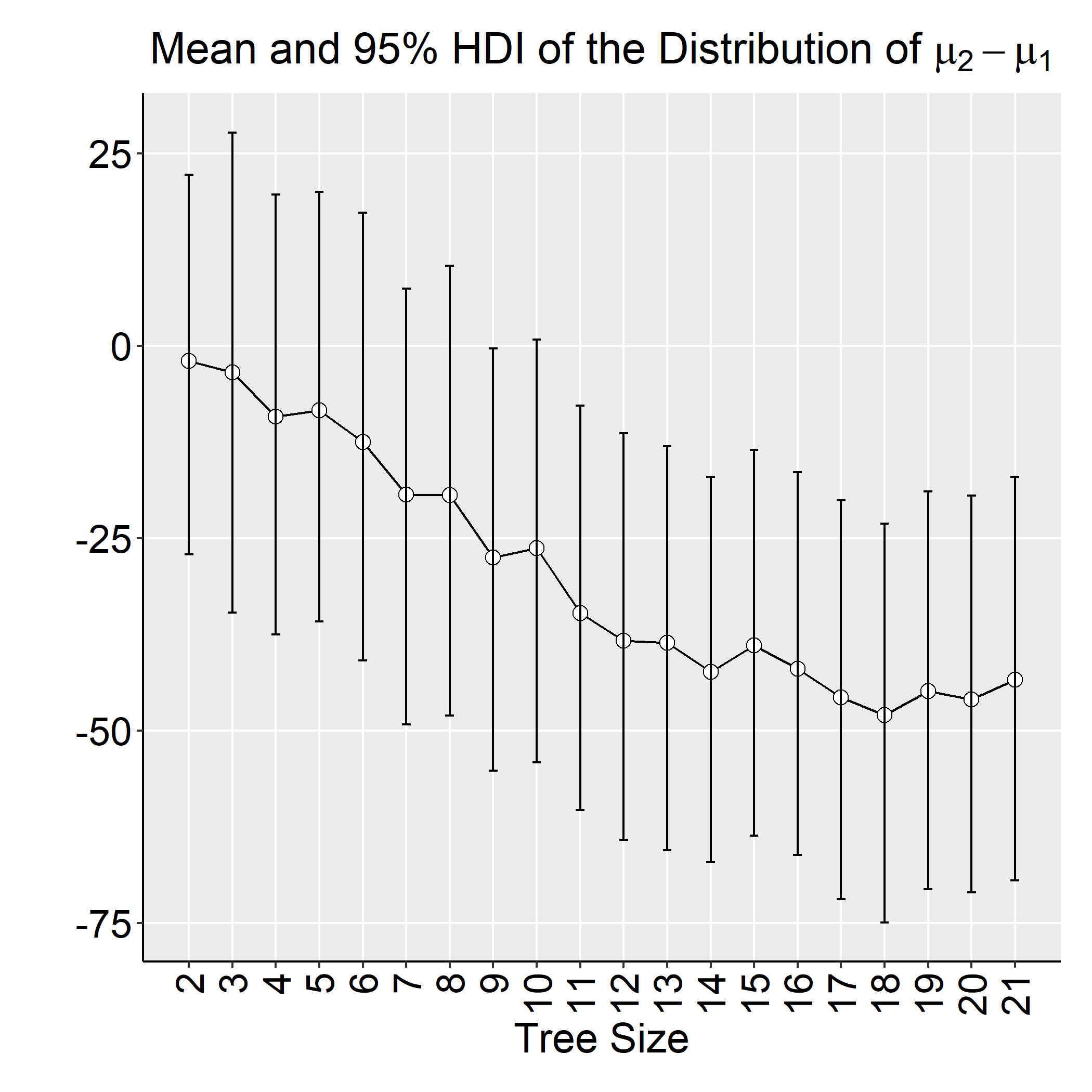} \\
			\small (d) Bike
		\end{tabular} \hspace{-8pt}
		\begin{tabular}[b]{c}
			\includegraphics[scale=.25]{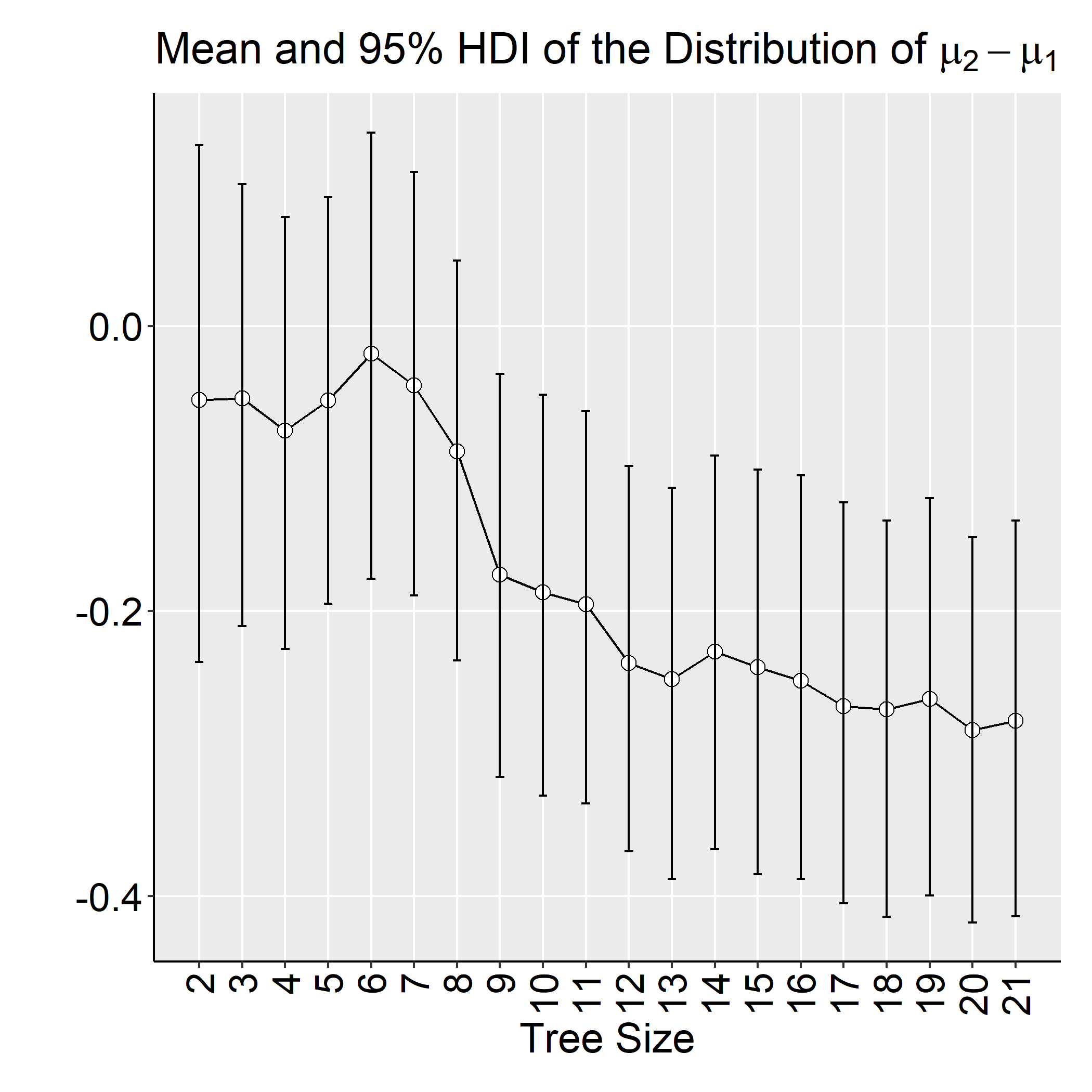} \\
			\small (e) Auto mpg
		\end{tabular}\hspace{-8pt}
		\begin{tabular}[b]{c}
			\includegraphics[scale=.25]{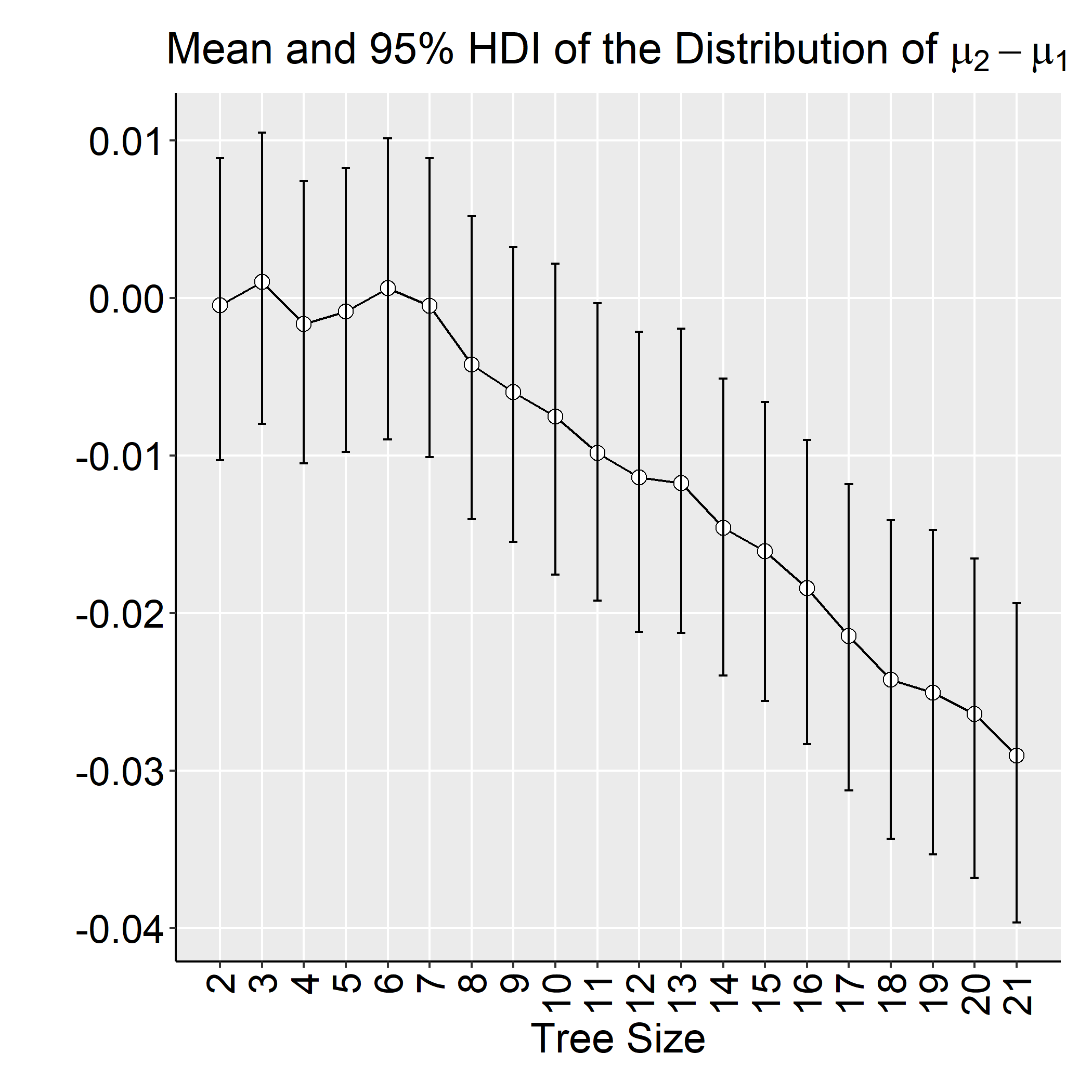} \\
			\small (f) Wine quality
		\end{tabular}
		\caption{Results of Bayesian t-test that shows the mean and $95\%$ highest density interval of the distribution of difference of means. $\mu_1$ and $\mu_2$ refer to the means of the distributions obtained for the interpretability prior and interpretability utility approach, respectively}
		\label{t-test}
	\end{figure}
	\subsubsection{Interpretability Prior vs Interpretability Utility}
	
	In this subsection, we compare our approach with the interpretability prior approach, in terms of the capability of the methods to trade off between accuracy and interpretabilityr. BART is used as the reference model, and CART is used as the interpretable model family. The interpretability prior approach fits a CART model directly to the training data where the prior assumption is that CART models are simple to interpret. On the other hand, our approach fits the CART model to the reference model, by optimization of the interpretability utility.
	
	Figure \ref{perf_comp} demonstrates the results using all the data sets introduced in Subsection \ref{subsec:data}. The results are averaged over $50$ runs. It can be seen that \textbf{the most accurate models with any level of complexity (interpretability) are obtained with our proposed approach}\footnote{The single exception happened in auto risk data set with tree size of $3$.}. 
	
	To test the significance of the differences in the results, we performed the Bayes $t$-test \cite{kruschke2013bayesian}. The approach works by building up a complete distributional information for the mean and standard deviation of each group\footnote{For each tree size, there are two groups of $50$ RMSE values: one for the interpretability prior approach, and one interpretability utility approach.} and constructing a probability distribution over their differences using MCMC estimation. From this distribution, the mean credible value as the best guess of the actual difference and the $95\%$ Highest Density Interval (HDI) as the range were the actual difference is with $95\%$ credibility are shown in Figure \ref{t-test}. When the $95\%$ HDI does not include zero, there is a credible difference between the two groups. As shown in the figure, for all data sets and for highly interpretable models (highly inaccurate), 
	the difference between the two approaches is not significant (HDI contains zero). This is expected since by increasing the interpretability, the ability of the interpretable model to explain variability of the data or of the reference model decreases, and both approaches provide almost equally poor performance. However, by increasing the complexity (equivalently decreasing interpretability) to a reasonable level, we see that the differences of the two approaches become significant for all data sets.
	\begin{table}[]
		\centering
		\caption{Comparison of our approach with two non-Bayesian counterparts: node harvest and BATrees. The RMSE values are shown in terms of $\mbox{mean}\pm\mbox{sd}$. Sizes are shown in terms of $\mbox{mean}\pm\mbox{sd}$ of number of leaf nodes for BATrees and number of nodes with non-zero coefficients for node harvest. For BATrees, the predictive performance of its reference model is shown in the parantheses.  For node harvest, it was not possible to obtain the performance of the reference model since the R package provides no means for that. Best RMSE values are bolded.}
		\begin{tabular}{c|c|c|c|c||c|c}
			\cline{2-7}
			\multirow{3}{*}{} & \multicolumn{4}{c||}{RMSE}  & \multicolumn{2}{c}{Size} \\ \cline{2-7} 
			& \multirow{2}{*}{Node Harvest} & \multirow{2}{*}{BATrees} & \multicolumn{2}{c||}{Our Approach} & \multirow{2}{*}{Node Harvest} & \multirow{2}{*}{BATrees} \\ \cline{4-5}
			&  &  & size = 10 & size = 15 &  &  \\ \hline
			
			\multirow{2}{*}{Body fat}  & \bm{$5.14\pm 0.37$}  & $5.26\pm 0.42$  & \multirow{2}{*}{$5.15\pm 0.36$} & \multirow{2}{*}{$5.16\pm 0.37$} & \multirow{2}{*}{$34.4 \pm 6.5$} & \multirow{2}{*}{$19.5 \pm 5.7$}  \\
			& * & ($4.84 \pm 0.38$) &   &   &   &   \\ \hline
			
			\multirow{2}{*}{Baseball}  & $783.1\pm 72.5$ & $1020.1\pm 69.9$ & \multirow{2}{*}{\bm{$755\pm 81.1$}} & \multirow{2}{*}{$768.6\pm 92.5$} & \multirow{2}{*}{$35.5 \pm 5.4$} & \multirow{2}{*}{$15.3 \pm 6.5$}  \\
			& * & ($755.9 \pm 67.8$) &   &   &   &   \\ \hline
			
			\multirow{2}{*}{Auto risk} & $0.78\pm 0.06$ & $0.79\pm 0.1$ & \multirow{2}{*}{$0.76\pm 0.1$} & \multirow{2}{*}{\bm{$0.75\pm 0.1$}} & \multirow{2}{*}{$37.2 \pm 12.4$} & \multirow{2}{*}{$20.4 \pm 4.8$}  \\
			& * & ($0.64 \pm 0.09$)  &   &   &  & \\ \hline
			
			\multirow{2}{*}{Bike} & $913.3\pm 67.2$ & $907.4\pm 71.5$ & \multirow{2}{*}{$934.4\pm 68.9$} & \multirow{2}{*}{\bm{$904.2\pm 60.6$}} & \multirow{2}{*}{$47.6 \pm 9.1$} & \multirow{2}{*}{$33.5 \pm 6.3$}  \\
			& * & ($681.8 \pm 60.7$)  &   &   &   &   \\ \hline
			
			\multirow{2}{*}{Auto mpg} & $3.47 \pm 0.33$ & \bm{$3.39 \pm 0.27$} & \multirow{2}{*}{$3.51 \pm 0.31$} & \multirow{2}{*}{$3.43 \pm 0.33$} & \multirow{2}{*}{$58.8 \pm 8.1$} & \multirow{2}{*}{$28 \pm 4.4$}  \\
			& * & ($2.83 \pm 0.34$) &   &   &   &   \\ \hline
			
			\multirow{2}{*}{Wine quality} & $0.67 \pm 0.03$ & \bm{$0.66 \pm 0.02$} & \multirow{2}{*}{$0.67 \pm 0.02$} & \multirow{2}{*}{\bm{$0.66 \pm 0.02$}} & \multirow{2}{*}{$51.3 \pm 7$} & \multirow{2}{*}{$43 \pm 13.2$}  \\
			& * & ($0.6 \pm 0.02$) &   &   &   &   \\ \hline
		\end{tabular}
		\label{T:comparison}
	\end{table}
	
	Finally, we further compared the performance of our proposed approach with two non-Bayesian counterparts, i.e., BATrees \cite{breiman1996born} and node harvest \cite{meinshausen2010node}. BATrees employs a single decision tree that best mimics the predictive behavior of a tree ensemble. Random forest is used as the reference model for BATrees. Node harvest simplifies a tree ensemble, i.e., random forest, by use of the shallow parts of the trees. We chose these approaches with random forest as their black-box model for the comparison for two reasons:
	\begin{itemize}
		\item[i.] to the best of our knowledge, there is no approach particularly established for explaining Bayesian tree ensemble models, i.e., BART. The approach of Hara and Hayashi \cite{hara2016making} can be modified for this objective; however, it requires inputs in terms of rules extracted from the tree ensemble, which calls for considerable of extra work.
		\item[ii.] BART can be considered as a Bayesian interpretation of random forest, and it has been revealed with some synthetic and real-data experiment that they have similar predictive performances \cite{hernandez2018bayesian}.
	\end{itemize}
	For node harvest, we used the R implementation with default setting. For BATrees, the Python implementation in \cite{hara2016making} is used with the depth of BATrees chosen from $\{3, 4, 5, 6\}$ using 5-fold cross validation. The measure of complexity for node harvest is the total number of nodes with non-zero coefficients.
	
	Table \ref{T:comparison} demonstrates the results. The results are averaged over $50$ runs with the same seed value used for the experiments in Figure \ref{perf_comp}. The table shows that our proposed approach attained much better trade-off between accuracy and interpretability compared to BATrees and node harvest. For $4$ data sets, our approach provides higher accuracies even with smaller sizes. For the rest, still our approach provides comparable predictive performance with a complexity of about half of the complexities of node harvest and BATrees. The differences between the bolded RMSE values with the rest of the RMSEs in Baseball, Auto risk and Wine quality data sets are significant using the Bayes $t$-test, while for other data sets the differences are not significant. According to the table, node harvest tends to generate more complex surrogate models. This is expected since in node harvest, the surrogate model is still an ensemble of shallow trees.
	
	\subsubsection{Stability Analysis}
	
	The goal of interpretable ML is to provide a comprehensive explanation of the predictive behavior of the black-box model to the decision maker. However, perturbation in the data or adding new samples may affect the learned interpretable model and lead to a very different explanation. This instability can cause problems for decision makers. Thereby, it is important to evaluate the stability of different interpretable ML approaches. 
	\begin{table}[!]
		\centering
		\caption{Bootstrap instability values in the form of $\mbox{mean}\pm \mbox{std}$. Best values are bolded.}
		\begin{tabular}{l|c|c|c|c|c|c}
			Interpretability  & Body fat & Baseball & Auto risk & Bike & Auto mpg & Wine quality\\
			\hline
			~~Prior & $0.71\pm 0.11$ & $0.84 \pm 0.08$ & $\pmb{0.79 \pm 0.19}$ & $\pmb{0.68 \pm 0.09}$ & $0.70 \pm 0.14$ & $0.74 \pm 0.11$ \\ 
			~~Utility &  $\pmb{0.62\pm 0.19}$ & $\pmb{0.83 \pm 0.07}$ & $0.81 \pm 0.16$ & $\pmb{0.68 \pm 0.09}$ & $\pmb{0.64 \pm 0.14}$ & $\pmb{0.70 \pm 0.13}$ \\ 
		\end{tabular}
		\label{T:stab.analys}
	\end{table}
	For this objective, we propose the following procedure for stability analysis of interpretable ML approach.
	
	Using a bootstrapping procedure with $10$ iterations, we compute pairwise dissimilarities of the interpretable models obtained using each approach and report the mean and standard deviation of the dissimilarity values as their instability measure (smaller is better). We used the dissimilarity measure proposed in \cite{briand2009similarity}. Assuming we are given two regression trees $T_1$ and $T_2$, for each internal node $t$, the similarity of the trees at node $t$ is computed by
	\begin{equation}
	S_{(1,2)}^t = I_{k=k'}^t\left(1 - \frac{\mid\!\delta_1^t - \delta_2^t\!\mid}{\mbox{range}(X_k)} \right)
	\end{equation}
	where $I_{k=k'}^t$ is the indicator that determines whether the feature used to grow node $t$ in $T_1$ is identical to the one used in $T_2$ ($I_{k=k'}^t=1$) or not, $\delta_1^t$ and $\delta_2^t$ are pivots used to grow the node $t$ in $T_1$ and $T_2$, respectively, and $\mbox{range}(X_k)$ is the range of values of feature $k$. Finally, the dissimilarity of the two decision trees is computed as $d\left(T_1,T_2\right)=1-\sum_{t\in{\mbox{internal\_nodes}}} q^t S_{(1,2)}^t$ where $q^t$ are user specified weight value which we set to $1/b$ where $b$ is the number of terminal nodes. The reported values are averaged over 45 values ($10$ bootstraping iterations result in $\left(10\times 9\right) / 2 = 45$ pairs of explainable models).
	
	Table \ref{T:stab.analys} compares the two approaches over the data sets introduced in Section \ref{subsec:data}. The interpretability utility approach generated on average more stable models for most data sets; however, drawing a general conclusion is not possible because except body fat, for the rest of the data sets, the differences are not significant according to the Bayes $t$-test.
	
	\subsection{Local Interpretation}\label{subsec:local_exp}
	
	We next demonstrate the ability of the proposed approach in locally interpreting the predictions of a Bayesian predictive model. BART \footnote{In this experiment, we set the number of trees to $50$ with nskip and ndpost set to $1000$ and $2000$ respectively, for faster run.} is used as the black box model and CART is used as the interpretable model family. For the CART model, we set the maximum depth of the decision trees to $3$ to obtain more interpretable local explanations. We compare with LIME\footnote{We use the `lime' package in R (\href{https://cran.r-project.org/web/packages/lime/lime.pdf}{https://cran.r-project.org/web/packages/lime/lime.pdf}) for the implementation.} which is a commonly used baseline for local interpretation approaches. Decision trees obtained by our approach to locally explain predictions of the BART model, used on average $2.03$ and $2.4$ features for the Boston housing and the auto risk data sets, respectively. Therefore, to maximize comparability, we set the feature selection approach of LIME to ridge regression and select the $2$ features with the highest absolute weights to be used in the explanation\footnote{MSEs of LIME with $3$ features are, respectively, $2.48$ and $0.006$ for Boston housing and Auto risk data sets.}.  We use the standard quantitative metric for local fidelity: $\mathbb{E}_x\left[\mbox{loss}\left(\mbox{interp}_x(\bm{x}), \mbox{pred}(\bm{x})\right)\right]$ where given a test data $\bm{x}$, $\mbox{interp}_x(x)$ refers to the prediction of the local interpretable model (fitted locally to the neighborhood of $\bm{x}$) for $\bm{x}$, and $\mbox{pred}(\bm{x})$ refers to the prediction of the black-box model for $\bm{x}$. We used locally weighted square loss as the loss function with $\pi_x=\mathcal{N}\left(\bm{x}, \sigma^2\bm{I}\right)$ where $\sigma = 1$.
	
\begin{figure}
		\centering
		\includegraphics[width = 0.5\textwidth]{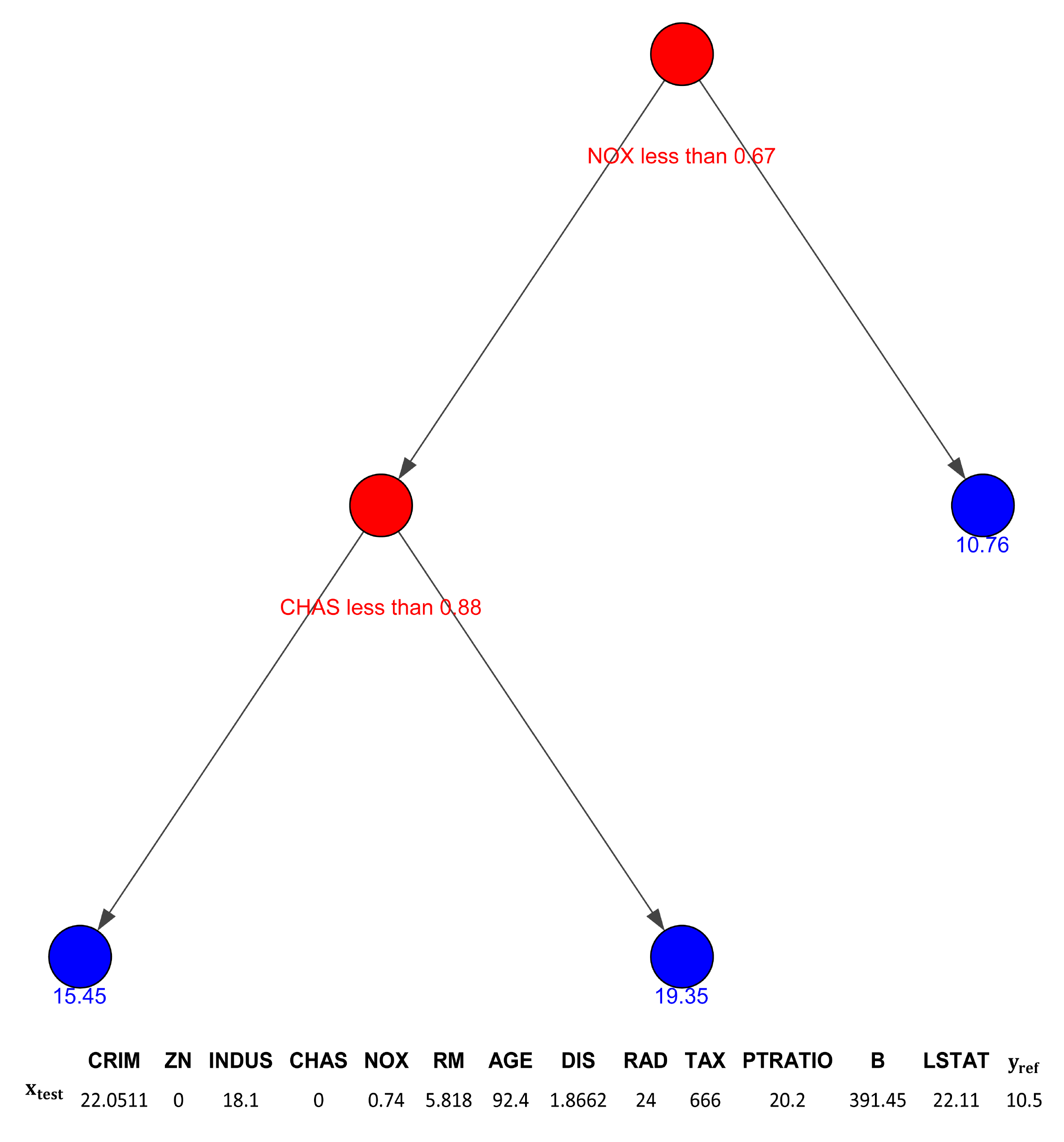}
		\caption{Example of a decision tree obtained by the interpretability utility approach to locally explain the prediction of the BART model ($y_{ref}$ is the mean of the predictions of the $2000$ posterior draws) for the particular test data $x_{test}$. Using only $2$ features, our approach predicts the output $10.76$. LIME with $2$ features predicts the output to be $14.06$, and with $3$ features, LIME prediction is $13.18$.}
		\label{example_tree}
	\end{figure}

	\begin{table}
		\caption{Comparison of the local fidelity of LIME and Interpertability utility when being used to explain predictions of BART. Best values are bolded.}
		\centering
		\begin{tabular}{l|cc}
			\multicolumn{1}{l|}{Dataset} & \multicolumn{1}{l}{LIME} & \multicolumn{1}{l}{Interpretability Utility} \\ \hline
			Boston housing & $4.86$ & $\bm{2.94}$\\
			Auto risk & $0.014$ & $\bm{0.010}$ \\
		\end{tabular}
		\label{local_fidelity_comp}
	\end{table}
	Each data set is divided into 90\%/10\% training/test split. For each test data, we draw $200$ samples from the neighborhood distribution. Table \ref{local_fidelity_comp} shows that our approach produces more accurate local explanation for both data sets. Figure \ref{example_tree} shows, as an example, a decision tree constructed by our proposed approach to locally explain the prediction of the BART model for the particular test data shown in the figure from Boston housing data set. It can be seen that using only two features, our proposed approach obtains good local fidelity while maintaining interpretability with a decision tree with only $3$ leaf nodes.
	
	\section{Conclusion}
	We presented a novel approach to construct interpretable explanations in the Bayesian framework by formulating the task as optimizing a utility function instead of changing the priors. This is obtained by first fitting a Bayesian predictive model which does not compromise accuracy, termed as a reference model, to the training data, and then project the information in the predictive distribution of the reference model to an interpretable model. The approach is model agnostic, implying that neither the reference model nor the interpretable model is restricted to a certain model. In the current implementation, the interpretable model, i.e., CART, is not a Bayesian predictive model; however, it is straightforward to extend the formulation to the case where a Bayesian predictive model, e.g., Bayesian CART \cite{denison1998bayesian}, is used as the interpretable model. This remains for future. The approach also allows accounting for model uncertainty in the explanations. Through experiments, we demonstrated that the proposed approach outperforms the alternative approach of restricting the prior, in terms of accuracy, interpretability and stability. Furthermore, we showed that the proposed approach performs comparable to non-Bayesian counterparts such as BATrees and node harvest even when they have higher complexities (equivalently less interpretability).
	
	\section{Acknowledgments}
	This work was supported by the Academy of Finland (Flagship programme: Finnish Center for Artificial Intelligence FCAI, grants 294238, 319264 and 313195), by the Vilho, Yrj\"{o} and Kalle V\"{a}is\"{a}l\"{a} Foundation of the Finnish Academy of Science and Letters, by the Foundation for Aalto University Science and Technology, and by the Finnish Foundation for Technology Promotion (Tekniikan Edist\"{a}miss\"{a}\"{a}ti\"{o}). We acknowledge the computational resources provided by the Aalto Science-IT Project.
	
	\bibliography{reference.bib}
	\bibliographystyle{plainnat}

\end{document}